%% file: main.tex
\documentclass{article}

\usepackage[main, final]{neurips_2026}

\usepackage[utf8]{inputenc} 
\usepackage[T1]{fontenc}    
\usepackage{hyperref}       
\usepackage{url}            
\usepackage{booktabs}       
\usepackage{amsfonts}       
\usepackage{nicefrac}       
\usepackage{microtype}      
\usepackage[dvipsnames,table,xcdraw,svgnames]{xcolor}
\definecolor{tabhighlight}{HTML}{EAEAEA} 
\definecolor{lightgreen}{HTML}{D9EAD3}   

\usepackage{natbib}
\usepackage{microtype}
\usepackage{graphicx}
\usepackage{subcaption}
\usepackage{booktabs} 
\usepackage{hyperref}
\usepackage{amsmath}
\usepackage{amssymb}
\usepackage{mathtools}
\usepackage[capitalize,noabbrev]{cleveref}

\newtheorem{Proposition}{\textbf{Proposition}}[section]

\newtheorem{Theorem}{\textbf{Theorem}}[section]
\newtheorem{Proof}{\textbf{Proof}}[section]
\newtheorem{Assumption}{\textbf{Assumption}}[section]
\usepackage[textsize=tiny]{todonotes}
\usepackage{algorithm}
\usepackage{algorithmic}
\usepackage{threeparttable}
\usepackage{multicol}
\usepackage{multirow}
\usepackage{wrapfig}

\title{Dirichlet-Guided Group Forecasting for Alleviating Over-smoothing in Time Series Forecasting}

\author{%
  \textbf{Xingyu Zhang}$^{1,2}$ \quad \textbf{Jingyao Wang}$^{1,2}$ \quad \textbf{Xin Yu}$^{1,2}$ \quad \textbf{Zeen Song}$^{3}$ \\
  \textbf{Jianqi Zhang}$^{1,2}$ \quad \textbf{Changwen Zheng}$^{1}$ \quad \textbf{Wenwen Qiang}$^{1}$\thanks{Corresponding author.} \\
  $^{1}$ National Key Laboratory of Space Integrated Information System, Institute of Software, \\
  Chinese Academy of Sciences, Beijing, China \\
  $^{2}$ University of Chinese Academy of Sciences, Beijing, China \\
  $^{3}$ Guizhou Power Grid Co., Ltd. Digital-Intelligent Operations Center,
  Guizhou, China \\
  \texttt{\{zhangxingyu23,wangjingyao22,yuxin261\}@mails.ucas.ac.cn,\ zeensong@qq.com} \\
  \texttt{jluzhangjianqi@163.com,\ \{changwen,qiangwenwen\}@iscas.ac.cn}
}

\begin{document}

\maketitle

\begin{abstract}
Time series forecasting often suffers from over-smoothing, especially when future dynamics are multi-modal. Forecasts may follow the coarse trend of the observed future, but fail to preserve sharp changes, oscillations, turning points, and regime transitions that define plausible dynamic evolution. In this work, we revisit over-smoothing from the perspective of latent dynamical mode compression: under partial observation and single-realization supervision, multiple plausible future modes can be weakened, merged, or averaged during forecasting. Based on this view, we propose Dirichlet-Guided Group Forecasting (DGF), a mode-preserving forecasting framework that explicitly models multiple mode-conditioned predictive distributions and uncertainty over their selection probabilities. DGF uses a Dirichlet-guided hierarchical sampling mechanism and reward-based optimization to encourage forecasts that are accurate, dynamically consistent, and mode-distinct. Extensive experiments on real-world forecasting benchmarks show that DGF reduces over-smoothing while improving forecasting accuracy, diversity, and dynamical consistency. Our code is publicly available at \url{https://github.com/Zh-XY22/DGF}.
\end{abstract}

\section{Introduction}
\begin{figure}
    \centering
    \includegraphics[width=0.95\linewidth]{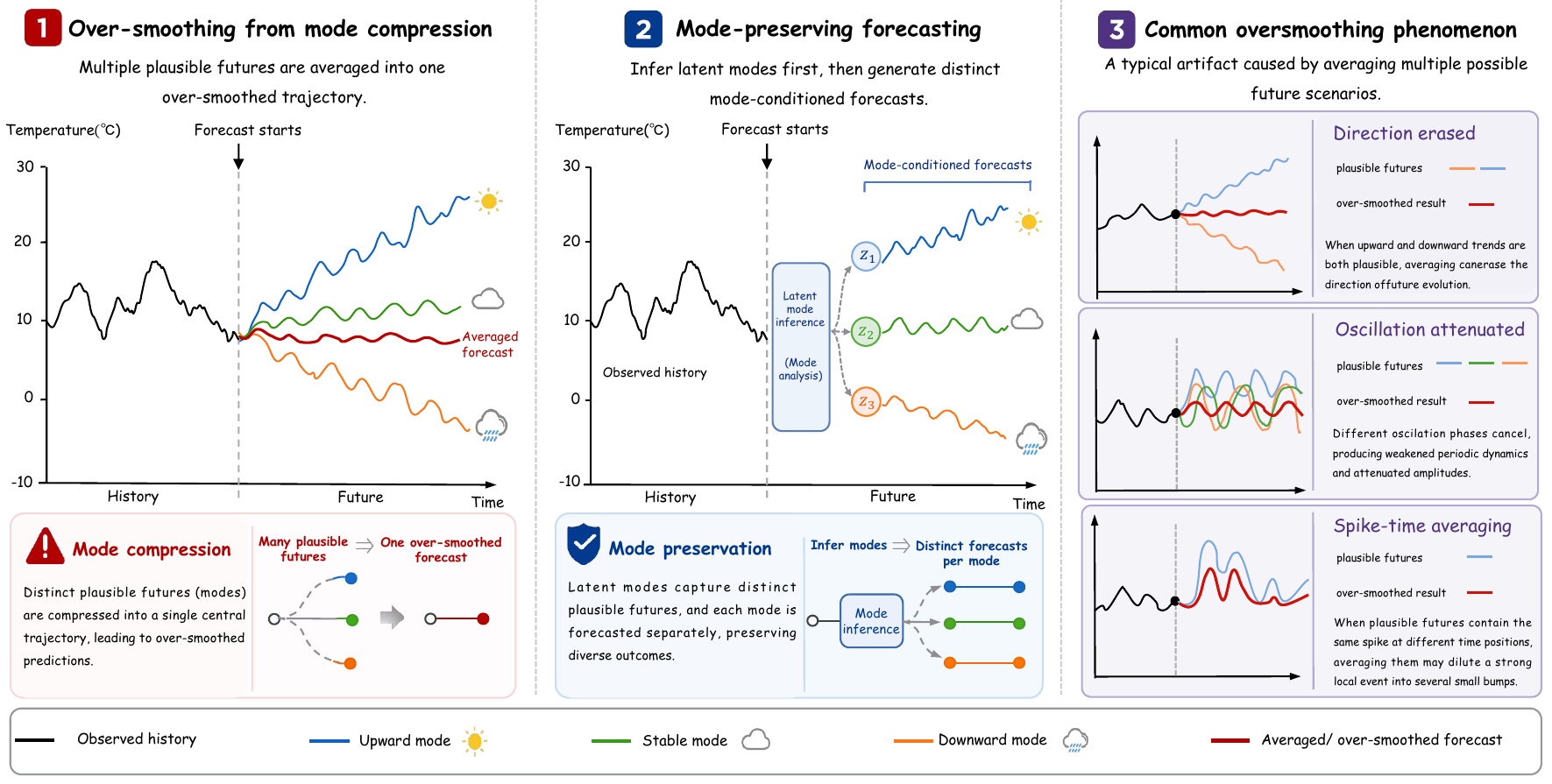}
    \caption{Illustration of over-smoothing and mode-preserving forecasting. Under the same observed history, multiple latent dynamical modes may lead to distinct plausible futures. A centralized forecast can compress these incompatible modes into an averaged trajectory with weakened dynamics, whereas mode-preserving forecasting maintains separate mode-conditioned trajectories.}
    \label{fig:oversmoothing}
\end{figure}

Time series forecasting (TSF) is a fundamental task in modern data-driven decision-making, with broad applications \citep{TSF2ECL, TSF2weather, TSF2traffic}. However, many real-world forecasting models still suffer from over-smoothing, especially in multi-step prediction: forecasts may capture the coarse trend but fail to reproduce sharp peaks, abrupt changes, local oscillations, frequency variations, turning points, and regime shifts \citep{HCAN}. As illustrated in Figure~\ref{fig:oversmoothing}, such forecasts may achieve acceptable average error, yet lose the dynamic structures that are critical for downstream decisions \citep{DILATE}.

We rethink over-smoothing as a problem of latent dynamical mode compression. In many real systems, the observed history is only a partial view of the underlying state, while future evolution may also depend on unobserved disturbances, external conditions, hidden states, or stochastic mechanisms \citep{rangapuram2018deep, salinas2020deepar}. Consequently, similar observed histories can lead to qualitatively different future patterns, such as rising, declining, oscillatory, abrupt-transition, or regime-switching trajectories \citep{DeepState, TimeGrad}. These patterns are not merely small perturbations around a common future; they may correspond to incompatible dynamic modes whose valid trajectory sets are not closed under linear averaging, a notion we formalize in Section~\ref{sec:problem_analysis}. However, each training input is usually paired with only one realized future trajectory, which reveals only one possible mode and leaves other dynamically plausible modes unobserved. Under such single-realization supervision, a forecasting model may suppress less frequently realized modes, merge nearby modes, or average incompatible modes into a trajectory that no longer preserves the structure of any individual future. From this perspective, over-smoothing is not only a loss-induced or frequency-domain artifact \citep{HCAN, DILATE, STRIPE}, but can also be understood as the compression of non-convex latent dynamical modes.

This perspective suggests that mitigating over-smoothing requires more than generating multiple future samples. If the samples are drawn from an implicit output distribution without exposing their latent mode structure, they may still fail to preserve distinct dynamical regimes. A mode-preserving forecaster should instead separate two questions: what mode-conditioned futures are possible, and how strongly the current history supports each mode. This requires explicitly maintaining multiple mode-conditioned predictive distributions and modeling uncertainty over their mode-selection probabilities \citep{DeepState, moiraimoe}. However, representation alone is not sufficient. Since each input is supervised by only one realized future, directly optimizing all candidate modes toward the same ground truth can collapse different mode-conditioned distributions back to the observed trajectory. Therefore, a mode-preserving forecaster also needs an optimization signal that keeps generated trajectories accurate while encouraging dynamical consistency and valid diversity.

To this end, we propose {Dirichlet-Guided Group Forecasting (DGF)}, a mode-preserving forecasting framework for mitigating over-smoothing. DGF learns multiple mode-conditioned predictive distributions to represent candidate future dynamics. It further places an input-dependent Dirichlet distribution over the mode-selection probability vector, separating the prediction of candidate futures from their selection. Forecasts are generated by sampling mode-selection probabilities and drawing trajectories from selected mode-conditioned distributions, avoiding deterministic aggregation of incompatible modes. To train this hierarchical stochastic process, DGF uses a GRPO-based objective \citep{deepseekmath} with trajectory-level accuracy and dynamical consistency rewards, group-level credit assignment, and a separate component-diversity regularizer. Experiments on real-world benchmarks demonstrate that DGF improves forecasting performance while producing more diverse and dynamically consistent future trajectories.
In summary, our main contributions are as follows:
\begin{itemize}
    \item We formulate over-smoothing in time series forecasting as latent dynamical mode compression under single-realization supervision, where multiple plausible future modes may be weakened, merged, or averaged in the learned forecasting process.
    
    \item We propose DGF, a mode-preserving probabilistic forecasting framework that jointly learns mode-conditioned predictive distributions and models second-order uncertainty over their mode-selection probability vector with a Dirichlet distribution.
    
    \item We develop a GRPO-based training objective that optimizes the hierarchical sampling process with accuracy, dynamical consistency, and diversity rewards.  
    
    \item Experiments on real-world forecasting benchmarks show that DGF reduces over-smoothing while improving forecasting accuracy, diversity, and dynamical consistency.
\end{itemize}

\section{Related Works}

\paragraph{Probabilistic and Generative Forecasting}
Probabilistic and generative forecasting methods aim to represent the uncertainty of future time series beyond a single deterministic trajectory. Instead of producing only a point forecast, these methods model the conditional predictive distribution of future states given historical observations. Early parametric approaches typically assume predefined likelihood families, such as Gaussian or Student's $t$ distributions. For example, DeepAR~\citep{salinas2020deepar} estimates likelihood parameters with autoregressive recurrent networks, while DeepState~\citep{rangapuram2018deep} combines state space models with deep neural networks to learn probabilistic latent states. To relax restrictive parametric assumptions, quantile-based and non-parametric methods such as MQ-RNN~\citep{wen2017multi} and SQF-RNN~\citep{gasthaus2019probabilistic} directly estimate multiple quantiles or spline-based quantile functions. More expressive generative forecasting methods further improve distributional flexibility: normalizing flow based methods~\citep{rasul2020multivariate} transform simple base distributions into complex predictive distributions, while diffusion-based approaches such as TimeGrad~\citep{TimeGrad} and CSDI~\citep{tashiro2021csdi} generate diverse future trajectories through iterative denoising. These methods provide important advances in representing non-Gaussian and multi-modal predictive uncertainty; however, they usually model uncertainty directly in the observation space, with latent dynamical modes and their selection structure remaining implicit. In contrast, we explicitly learn multiple mode-conditioned predictive distributions and use a Dirichlet-guided group mechanism to model and optimize uncertainty over their mode composition, thereby targeting over-smoothing from the perspective of latent mode compression.

\paragraph{Deep Learning for Time Series Forecasting}
Deep learning has substantially reshaped time series forecasting (TSF) by enabling flexible representation learning from complex temporal data~\citep{lecun_deep_2015,lim2021time}. Early neural forecasting methods mainly relied on recurrent neural networks to model sequential dependencies~\citep{how_RNN2TSF,tang2021building}, while convolutional architectures were later adopted to capture local temporal patterns through causal or dilated convolutions~\citep{bai2018empirical,zhan2023differential}. With the success of attention mechanisms, Transformer-based models became widely used for capturing long-range dependencies in TSF~\citep{TransformersinTS}. 
Recent studies have also revisited simpler yet effective architectures, including linear and MLP-based models such as DLinear~\citep{DLinear} and TiDE~\citep{TiDE}, as well as models with stronger temporal inductive biases, such as PatchTST~\citep{PatchTST}, TimesNet~\citep{TimesNet}, FEDformer~\citep{FEDformer}, FreDF~\citep{zhang2024not}, and iTransformer~\citep{iTransformer}. These methods have significantly improved forecasting accuracy and temporal representation learning, but they mainly focus on how to encode historical dependencies, while the preservation and optimization of multiple plausible future modes remains less directly addressed.

\paragraph{Large Language Models and Foundation Models for TSF}
Inspired by the strong generalization ability of large language models (LLMs), recent studies have explored adapting pre-trained language models to time series forecasting by reformulating numerical sequences as textual prompts or token sequences. PromptCast~\citep{PromptCast} converts time series into sentence-like prompts, LLMTime~\citep{LLMTime} represents numerical sequences as strings for zero-shot forecasting, and AutoTimes~\citep{AutoTimes} leverages autoregressive generation by conditioning LLMs on historical sequences. Other works align time series with the latent space of pre-trained models through reprogramming, prompt tuning, or lightweight fine-tuning, such as FPT~\citep{FPT}, Time-LLM~\citep{Time-LLM}, TEMPO~\citep{tempo}, and UniTime~\citep{UniTime}. Beyond adapting general-purpose LLMs, recent time-series foundation models are trained directly on large-scale temporal corpora to support zero-shot or few-shot forecasting across domains, including decoder-only or tokenized forecasting models such as TimeGPT~\citep{TimeGPT}, TimesFM~\citep{TimesFM}, and Chronos~\citep{Chronos}, as well as unified or patch-based foundation models such as MOMENT~\citep{MOMENT}, MOIRAI~\citep{Moirai}, Timer~\citep{Timer}, and Timer-XL~\citep{liu2025timerxl}. These works mainly focus on representation transfer, scaling behavior, long-context modeling, or cross-domain generalization.

\section{Preliminaries}

\subsection{Time Series Forecasting}
Let $X=[x_1,\dots,x_T]\in\mathbb{R}^{T}$ denote a historical series window, and let $Y=[x_{T+1},\dots,x_{T+S}]\in\mathbb{R}^{S}$ denote the future trajectory over a prediction horizon $S$. Given a training dataset $\mathcal{D}=\{(X^{(n)},Y^{(n)})\}_{n=1}^{N}$, time series forecasting aims to infer the future trajectory from the observed history. A common deterministic formulation learns a single-valued forecasting function $f_\theta$ and outputs $\hat{Y}=f_\theta(X)$. The model is typically trained by empirical risk minimization:
\begin{equation}
    \min_{\theta}\mathcal{L}(\theta) = \frac{1}{N} \sum_{n=1}^{N} \ell\left(f_\theta(X^{(n)}),Y^{(n)}\right),    
\end{equation}
where $\ell(\cdot,\cdot)$ measures the discrepancy between the predicted trajectory and the observed future. In contrast, probabilistic forecasting represents a conditional predictive distribution $p_\theta(Y\mid X)$, from which forecast samples or summary statistics can be obtained.

\subsection{The Dirichlet Distribution}
The Dirichlet distribution is a distribution over probability vectors on the simplex. Let $\boldsymbol{\pi}=[\pi_1,\dots,\pi_K]\in\Delta^{K-1}$, where $\pi_k\geq 0$ and $\sum_{k=1}^{K}\pi_k=1$. Given concentration parameters $\boldsymbol{\alpha}=[\alpha_1,\dots,\alpha_K]$ with $\alpha_k>0$, the Dirichlet density is
\begin{equation}
    {\rm Dir}(\boldsymbol{\pi}\mid\boldsymbol{\alpha}) = \frac{1}{B(\boldsymbol{\alpha})} \prod_{k=1}^{K} \pi_k^{\alpha_k-1},  \boldsymbol{\pi}\in\Delta^{K-1}, \text{where }  B(\boldsymbol{\alpha})=\frac{\prod_{k=1}^{K}\Gamma(\alpha_k)}{\Gamma(\sum_{k=1}^{K}\alpha_k)}.
    \label{eq:dirichlet}
\end{equation}
Let $\alpha_0=\sum_{k=1}^{K}\alpha_k$. The expected probability of component $k$ is $\mathbb{E}[\pi_k]=\alpha_k/\alpha_0$. The normalized parameters determine the mean probability vector, while the total concentration $\alpha_0$ controls dispersion around this mean. Thus, Dirichlet distributions with the same mean can express different levels of variability in mode-selection probabilities. In DGF, $\pi$ denotes a sampled mode-selection probability vector, and $\alpha$ parameterizes its input-dependent distribution.

\section{Problem Analysis and Motivation}\label{sec:problem_analysis}

\subsection{Over-Smoothing as Dynamic Structure Degradation}
Over-smoothing in TSF refers to the degradation of dynamic structures in predicted future trajectories. A smoothed forecast may follow the coarse trend of the observed future, yet suppress local peaks, abrupt changes, oscillations, or regime transitions. Such forecasts can be acceptable under average value-level metrics while being dynamically uninformative. Therefore, over-smoothing is better understood as a mismatch between value-level accuracy and dynamic-structure preservation.

Let $\hat{Y}$ be a predicted trajectory and $Y_{gt}$ be the observed future. We denote a conventional value-level discrepancy by $d_{\mathrm{val}}(\hat{Y},Y_{gt})$. To describe dynamic preservation, let $\Phi:\mathcal{Y}\rightarrow\mathcal{Z}$ be a dynamic feature mapping that summarizes trajectory-level properties such as trend, volatility, frequency pattern, turning points, and transition behavior. The corresponding dynamic-structure discrepancy is $d_{\mathrm{dyn}}(\hat{Y},Y_{gt})= D(\Phi(\hat{Y}),\Phi(Y_{gt}))$, where $D(\cdot,\cdot)$ measures discrepancy in the dynamic feature space. 
The key observation is $d_{\mathrm{val}}(\hat{Y},Y_{gt}) \ \text{small} \not\Rightarrow d_{\mathrm{dyn}}(\hat{Y},Y_{gt}) \ \text{small}$. Thus, a forecast may be close to the observed future in value space while losing the structural properties that make the trajectory dynamically plausible. This motivates analyzing over-smoothing through dynamic-structure preservation rather than only through average prediction error.

\subsection{Latent-Mode Structure of Future Uncertainty}
Future uncertainty in time series is often structured rather than purely random. For a given historical context $X$, the future trajectory $Y$ may depend on an unobserved discrete latent mode $M$, which represents possible dynamical regimes such as trend directions, volatility states, oscillatory patterns, abrupt changes, or transition behaviors. We use $M$ to describe future-relevant structure that is partially informed by the observed history, but not fully determined by it.
\begin{Assumption}[Partially observed latent future modes]\label{assumption}
For a historical context $X$, future trajectory $Y$, and discrete latent dynamical mode $M$, the mode $M$ satisfies
\begin{equation}
    I(X;M)>0, H(M\mid X)>0, I(M;Y\mid X)>0.    
\end{equation}
\end{Assumption}
This assumption states that the historical context contains information about the future mode, but does not determine it completely. Meanwhile, the latent mode provides additional information about the future trajectory beyond $X$ alone. Thus, $M$ is neither arbitrary noise independent of the input nor a redundant variable absorbed by the observed history; it captures partially observed future structure. Under this assumption, the conditional future distribution can be decomposed as
\begin{equation}
    p(Y\mid X) = \sum_{m=1}^{K}p(Y\mid X,M=m)p(M=m\mid X).
\end{equation}
Here, $p(Y\mid X,M=m)$ denotes the mode-conditioned future distribution, while $p(M=m\mid X)$ denotes the probability that the future follows the $m$-th latent mode under the observed history. This decomposition separates within-mode uncertainty, i.e., variability of trajectories under a fixed mode, from mode-selection uncertainty, i.e., uncertainty over which latent mode will be realized after conditioning on $X$. The key training difficulty is that the latent mode $M$ is not observed. For each historical window, the dataset usually provides only one realized future trajectory, which corresponds to one realized mode. Other dynamically plausible modes under the same observed history are not directly supervised. Therefore, even if the conditional distribution is multi-modal, the available supervision for each input is a single realization. This single-realization supervision is the setting in which mode-conditioned structures can be compressed or collapsed during training.

\subsection{Mode Averaging and Non-Convex Dynamic Structures}
The latent-mode decomposition shows that the conditional future distribution may contain multiple mode-conditioned trajectory structures. However, estimating a predictive distribution and producing forecast outputs are different problems. A predictive distribution describes possible futures, whereas a forecasting system must eventually output either a trajectory or a finite set of trajectories. If the output decision reduces incompatible modes to a centralized forecast, the resulting trajectory may lose the dynamic structure of every individual mode.

To formalize this issue, let $\mathcal{Y}\subseteq\mathbb{R}^{S}$ denote the trajectory space over the prediction horizon. For each latent mode $m$, let $\mathcal{S}_m\subseteq\mathcal{Y}$ denote the set of dynamically valid trajectories under that mode, and let $\mathcal{S}=\bigcup_{m=1}^{K}\mathcal{S}_m$ denote the union of all mode-specific valid trajectory sets. Such a set is generally non-convex: averaging an increasing and a decreasing trajectory may produce a flat trajectory, and averaging oscillatory trajectories with different phases may suppress their amplitudes. Therefore, a forecast obtained by averaging across modes may no longer belong to any valid mode-specific structure. The following theorem connects this geometric observation to point-risk minimization.

\begin{Theorem}[Point-risk minimization can induce dynamically invalid forecasts]
\label{thm:point_risk_invalid}
Assume that the dynamically valid trajectory set $\mathcal{S}=\bigcup_{m=1}^{K}\mathcal{S}_m$ is non-convex. Then there exists a conditional distribution $p(Y\mid X=x)$ supported entirely on $\mathcal{S}$ such that the Bayes-optimal predictor under squared loss, 
\begin{equation}
    f^\star(x)=\arg\min_{f(x)}\mathbb{E}[\|f(x)-Y\|_2^2\mid X=x],    
\end{equation}
is dynamically invalid
\begin{equation}
    Y\sim p(Y\mid X=x)\Rightarrow Y\in\mathcal{S} \text{almost surely},f^\star(x)\notin\mathcal{S}.    
\end{equation}
\end{Theorem}

The proof is provided in Appendix~\ref{app:proof_point_risk_invalid}. Theorem~\ref{thm:point_risk_invalid} shows that over-smoothing can arise from the interaction between non-convex dynamic mode structures and centralized point decisions. The possible futures may all be dynamically valid, but the squared-loss optimal point forecast can still lie outside the valid dynamic-mode set. In this case, the forecast is not merely inaccurate in value space; it is dynamically degraded because it compresses incompatible latent modes into a trajectory that preserves none of them. However, simply increasing the number of prediction heads does not remove this compression. Under single-realization supervision, if all heads are optimized against the same observed future without mode-specific assignment or separation, the heads receive identical supervision and can collapse to the same solution.

\begin{Proposition}[Naive multi-head supervision admits a collapsed optimum]
\label{prop:head_collapse}
Consider a multi-head forecaster with $K$ deterministic heads $\{\mu_k(X)\}_{k=1}^{K}$ trained on a single realized target $Y_{gt}$ for each input $X$ using $\mathcal{L}_{\mathrm{naive}}(X,Y_{gt}) = \sum_{k=1}^{K} \|\mu_k(X)-Y_{gt}\|_2^2$.
For any fixed training sample $(X,Y_{gt})$, the unique minimizer satisfies $\mu_1(X)=\mu_2(X)=\cdots=\mu_K(X)=Y_{gt}$. For the population risk under squared loss, the minimizer satisfies $\mu_1^\star(X)=\mu_2^\star(X)=\cdots=\mu_K^\star(X)=\mathbb{E}[Y\mid X].$
\end{Proposition}

The proof is provided in Appendix~\ref{app:proof_head_collapse}. Proposition~\ref{prop:head_collapse} explains why multiple predictive heads alone are insufficient for mode-preserving forecasting. If every head is directly pulled toward the same realized trajectory, then the model has no incentive to assign different heads to different latent modes. This is precisely the collapse mechanism that a mode-preserving method must counteract.

Together, Theorem~\ref{thm:point_risk_invalid} and Proposition~\ref{prop:head_collapse} identify two potential sources of mode compression: averaging incompatible futures at prediction time and training all predictive heads toward the same target. These observations motivate separating component prediction from component selection, together with training signals that encourage dynamic consistency and component separation. DGF implements this design through mode-conditioned predictive distributions, Dirichlet-guided group sampling, and hierarchical reward-based optimization.

\begin{figure*}
    \centering
    \includegraphics[width=0.95\linewidth]{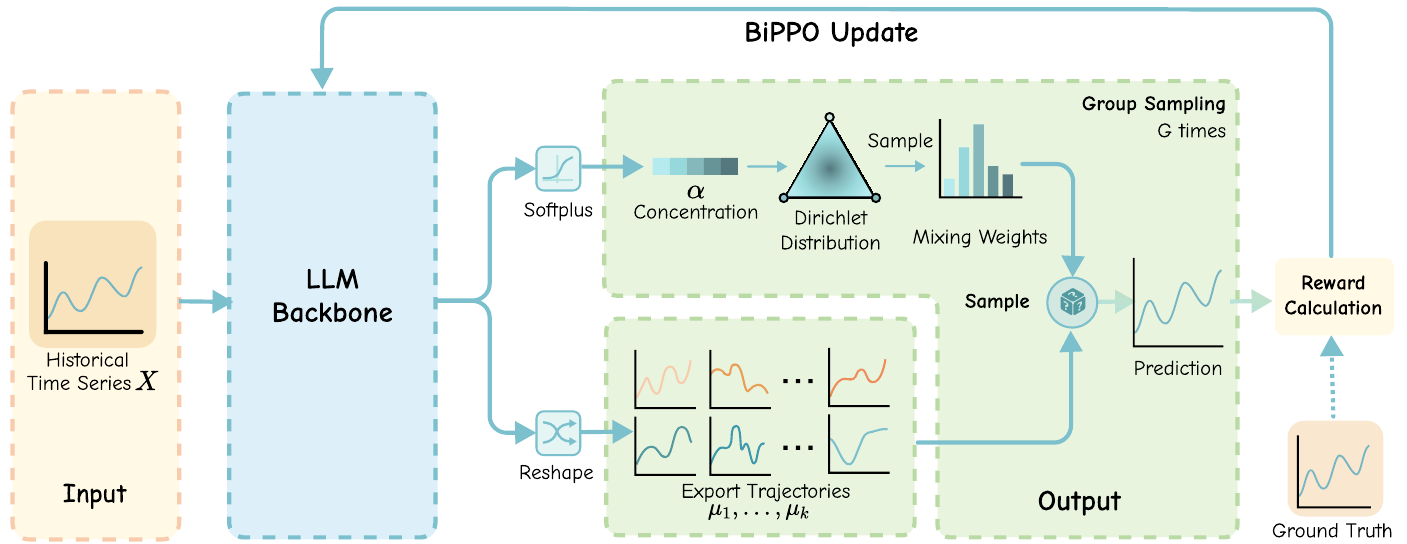}
    \caption{Overview of DGF. The backbone produces a contextual representation used to parameterize predictive components and a Dirichlet distribution over component-selection probabilities. For each input, $B$ probability vectors are sampled. Within each mini-group, one sampled vector is held fixed while $G$ forecasts are generated by sampling component indices and their corresponding predictive distributions. Trajectory-level and group-level credit assignment optimize the sampling process, while a separate diversity regularizer encourages component separation.}
    \label{fig:framework}
\end{figure*}

\section{Method}
\label{sec:method}

\paragraph{Overview.}
As shown in Figure~\ref{fig:framework}, DGF learns $K$ mode-conditioned predictive distributions and an input-dependent Dirichlet distribution over their selection probabilities. A sampled probability vector is shared by multiple forecasts, forming a mini-group. Hierarchical GRPO assigns credit to individual forecasts and sampled probability vectors, while a separate diversity regularizer encourages separation among predictive components.

\subsection{Dirichlet-Guided Group Forecasting}
To preserve latent future modes, DGF separates mode-conditioned prediction from mode selection.  Given a historical input $X$, an encoder first extracts a contextual representation $h=F_{\mathrm{enc}}(X)$. Based on $h$, the model produces $K$ mode-conditioned predictive distributions $q_{\theta,k}(Y\mid X)=g_{\theta,k}(h), k=1,\dots,K$, where each distribution is intended to capture one candidate latent dynamical mode. The distribution family can be chosen according to the task, such as Gaussian, Student-$t$, or other probabilistic output heads. In parallel, DGF predicts a Dirichlet concentration vector $\alpha_\theta(X)=[\alpha_{\theta,1}(X),\dots,\alpha_{\theta,K}(X)], \alpha_{\theta,k}(X)>0$. In practice, we parameterize it by evidence scores $e_\theta(X)=g_\alpha(h)$ and set $\alpha_{\theta,k}(X)=\alpha_{min} +\mathrm{softplus}(e_{\theta,k}(X))$. A deterministic softmax gate outputs one mode-selection probability vector for each input. DGF instead parameterizes a distribution over such vectors, $p(\pi\mid X)=\mathrm{Dir}(\pi\mid\alpha_\theta(X))$. The normalized parameters specify the expected component probabilities, while the total concentration controls variability around them. Sampled probability vectors represent alternative mode-composition hypotheses, which are evaluated through the group-level mechanism below.

For each input $X$, DGF samples $B$ independent mode-selection probability vectors,
$\pi_b\sim\mathrm{Dir}(\alpha_\theta(X))$, where $b=1,\ldots,B$.
Each vector is shared by a mini-group of $G$ forecasts. For the $i$-th forecast in group $b$, we sample
\begin{equation}
    z_{b,i}\sim\mathrm{Cat}(\pi_b), \qquad \hat{Y}_{b,i}\sim q_{\theta,z_{b,i}}(Y\mid X), \qquad i=1,\ldots,G.
\end{equation}
Here, $z_{b,i}$ is the selected predictive-component index. The joint density for a probability vector, component index, and forecast is
\begin{equation}
    p_\theta(\pi_b,z_{b,i},\hat{Y}_{b,i}\mid X) = \mathrm{Dir}(\pi_b\mid\alpha_\theta(X)) \,\mathrm{Cat}(z_{b,i}\mid\pi_b) \,q_{\theta,z_{b,i}}(\hat{Y}_{b,i}\mid X).
\end{equation}
If each trajectory independently samples its own Dirichlet vector, its marginal component-selection probability is
\begin{equation}
    P(z_i=k\mid X) = \mathbb{E}[\pi_{i,k}] = \frac{\alpha_{\theta,k}(X)}{\alpha_0(X)}, \qquad \alpha_0(X)=\sum_{k=1}^{K}\alpha_{\theta,k}(X).
\end{equation}
Group sharing preserves this single-trajectory marginal but induces dependence among component selections within a mini-group. It also allows each sampled composition to be evaluated using multiple forecasts. These mini-groups are credit-assignment units for the same input, distinct from optimization mini-batches.

\subsection{Reward Design and Hierarchical GRPO}
The hierarchical sampling process in DGF involves three different optimization targets. Sampled trajectories should be accurate and dynamically plausible; the Dirichlet sampler should favor mode compositions that produce high-quality forecast groups; and the $K$ mode-conditioned predictive distributions should remain separated to avoid mode collapse. Therefore, we separate trajectory-level rewards, Dirichlet-level credit assignment, and distribution-level diversity regularization.

For each sampled trajectory $\hat{Y}_{b,i}$ in the $b$-th mini-group, we define the trajectory reward as $r^{\mathrm{traj}}_{b,i}=\lambda_1 r_{\mathrm{acc},b,i}+\lambda_2 r_{\mathrm{dyn},b,i}$. The accuracy term grounds the forecast in the realized future, e.g., $r_{\mathrm{acc},b,i}=-\mathrm{MSE}(\hat{Y}_{b,i},Y_{\mathrm{gt}})$, or other forecasting scores such as MAE, CRPS, or log-likelihood. The dynamics term uses a fixed set of temporal descriptors 
\begin{equation}
\begin{aligned}
    r_{\mathrm{dyn},b,i}= &-\eta_1\|\Delta \hat{Y}_{b,i}-\Delta Y_{\mathrm{gt}}\|_1 -\eta_2\|\Delta^2 \hat{Y}_{b,i}-\Delta^2 Y_{\mathrm{gt}}\|_1 \\
    &-\eta_3\|A(\hat{Y}_{b,i})-A(Y_{\mathrm{gt}})\|_1 -\eta_4 d_{\mathrm{trans}}(X,\hat{Y}_{b,i}),
\end{aligned}
\end{equation}
where $\Delta$ and $\Delta^2$ capture local slope and curvature, $A(\cdot)$ is the normalized FFT amplitude spectrum, and $d_{\mathrm{trans}}$ penalizes discontinuity between the historical window and the predicted horizon. Thus, the trajectory reward evaluates both value-level accuracy and dynamic consistency.

We encourage diversity among the $K$ predictive components through a separate regularizer. Distances between sampled trajectories alone do not establish component separation: even identical predictive distributions can generate different samples. Moreover, a diversity bonus shared by all trajectories in a mini-group is removed by within-group advantage centering. We therefore regularize the predictive components directly:
\begin{equation}
    \mathcal{R}_{\mathrm{div}}^{K}(X)=\frac{1}{K(K-1)}\sum_{k\neq l}\min(D_{\mathrm{mode}}(q_{\theta,k}(\cdot\mid X),q_{\theta,l}(\cdot\mid X)),m),    
\end{equation}
Here, $D_{\mathrm{mode}}$ measures separation between predictive components, and the margin $m$ caps the reward for each pair. Let $\mu_k(X)$ denote the mean trajectory of $q_{\theta,k}(Y\mid X)$. We define component separation using value and dynamic-feature distances:
\begin{equation}
    D_{\mathrm{mode}}(q_{\theta,k},q_{\theta,l}) = D_{\mathrm{val}}(\mu_k,\mu_l) + \rho D_{\mathrm{dyn}}(\Phi(\mu_k),\Phi(\mu_l)).
\end{equation}
This regularizer encourages distinct component means and dynamic features. Once a pairwise distance reaches $m$, further separation yields no additional diversity reward. We optimize the hierarchical sampling process using separate trajectory-level and Dirichlet-level credit assignment. For sample-level credit assignment, forecasts are compared within the same mini-group. Given trajectory rewards $\{r^{\mathrm{traj}}_{b,i}\}_{i=1}^{G}$, we compute $\mu_b=\frac{1}{G}\sum_{j=1}^{G}r^{\mathrm{traj}}_{b,j}$, $\sigma_b=\mathrm{Std}(\{r^{\mathrm{traj}}_{b,j}\}_{j=1}^{G})$, and $A^{\mathrm{sample}}_{b,i}=\frac{r^{\mathrm{traj}}_{b,i}-\mu_b}{\sigma_b+\epsilon}$. The sample-level objective is 
\begin{equation}
    \mathcal{J}_{\mathrm{sample}}=\sum_b\sum_{i=1}^{G}A^{\mathrm{sample}}_{b,i}[\log p(z_{b,i}\mid\pi_b)+\log q_{\theta,z_{b,i}}(\hat{Y}_{b,i}\mid X)].    
\end{equation}
This term reinforces better trajectories and updates the selected mode-conditioned predictive distribution. For Dirichlet-level credit assignment, $\pi_b$ must be evaluated by the quality of the whole mini-group. Directly applying normalized sample-level advantages to $\log p_\theta(\pi_b\mid\alpha_\theta(X))$ would cancel out because $\pi_b$ is shared within the group. We therefore define the mini-group reward as $R_b=\frac{1}{G}\sum_{i=1}^{G}r^{\mathrm{traj}}_{b,i}$. Across $B$ mini-groups, we compute $\mu_R=\frac{1}{B}\sum_{b=1}^{B}R_b$, $\sigma_R=\mathrm{Std}(\{R_b\}_{b=1}^{B})$, and $A^{\mathrm{dir}}_b=\frac{R_b-\mu_R}{\sigma_R+\epsilon}$. The Dirichlet-level objective is 
\begin{equation}
    \mathcal{J}_{\mathrm{dir}}=\sum_{b=1}^{B}A^{\mathrm{dir}}_b\log p_\theta(\pi_b\mid\alpha_\theta(X)),
\end{equation}
which favors mode-composition hypotheses that generate better forecast groups. The final objective is $\mathcal{J}_{\mathrm{DGF}}=\mathcal{J}_{\mathrm{sample}}+\gamma\mathcal{J}_{\mathrm{dir}}+\lambda_3\mathcal{R}_{\mathrm{div}}^{K}$. Here, $\mathcal{J}_{\mathrm{sample}}$ optimizes individual trajectories and selected predictive heads, $\mathcal{J}_{\mathrm{dir}}$ optimizes the Dirichlet mode-composition distribution, and $\mathcal{R}_{\mathrm{div}}^{K}$ encourages separation among the $K$ predictive components.

\paragraph{Point forecasting.}
For deterministic LSF evaluation, we use the mean trajectory of the component with the largest expected selection probability:
\begin{equation}
    \bar{\pi}_k(X) = \frac{\alpha_{\theta,k}(X)}{\alpha_0(X)}, \qquad k^*=\arg\max_k\bar{\pi}_k(X), \qquad \hat{Y}_{\mathrm{point}}=\mu_{k^*}(X).
\end{equation}
We refer to this rule as MAP-mode output. It depends only on the input and model predictions and avoids averaging across component means. All main LSF results use this rule; alternative decision protocols are examined in Appendix F.

\section{Experiment}\label{exp}
We evaluate DGF on deterministic and probabilistic forecasting benchmarks. The experiments examine whether DGF preserves standard forecasting accuracy, improves probabilistic forecast quality, and mitigates over-smoothing and dynamic-structure degradation. Dataset details, finetuning baselines, metric definitions, full results, and component ablations are provided in Appendix~\ref{sec:app_imple_details} and \ref{app_ablation}.

\subsection{Main Results}
\paragraph{Long Sequence Forecasting}
We evaluate deterministic forecasting performance on standard LSF benchmarks following \citep{Moirai}. Since LSF evaluation requires a single reported trajectory, DGF reports the mode-selected forecast; alternative decision protocols are analyzed in Appendix~\ref{app_ablation}. As shown in Table~\ref{tab:avg_tsf_result}, DGF consistently improves both Moirai-small and Moirai-base across datasets and achieves the best average performance on most benchmarks. These results indicate that the proposed mode-preserving generation mechanism improves standard point-forecasting accuracy without relying on deterministic aggregation of all mode-conditioned distributions.

\begin{table*}[ht]
  \centering
  \caption{
Long sequence forecasting results averaged over prediction lengths \(\{96,192,336,720\}\). The best results for each metric are highlighted in \textbf{bold}.
}  
  \label{tab:avg_tsf_result}%
  \resizebox{0.95\textwidth}{!}{
\begin{tabular}{lcccccccccccc}
\toprule
\multirow{2}[2]{*}{\centering Method} &  \multicolumn{2}{c}{\textbf{ETTm1}} & \multicolumn{2}{c}{\textbf{ETTm2}} & \multicolumn{2}{c}{\textbf{ETTh1}} & \multicolumn{2}{c}{\textbf{ETTh2}} & \multicolumn{2}{c}{\textbf{Electricity}} & \multicolumn{2}{c}{\textbf{Weather}} \\
\cmidrule(lr){2-3}  \cmidrule(lr){4-5}  \cmidrule(lr){6-7}  \cmidrule(lr){8-9}  \cmidrule(lr){10-11}  \cmidrule(lr){12-13}
& MSE & MAE & MSE & MAE & MSE & MAE & MSE & MAE & MSE & MAE & MSE & MAE \\
\midrule
{DLinear\citeyearpar{DLinear}} 
& 0.403 & 0.407 & 0.350 & 0.401 & 0.456 & 0.452 & 0.559 & 0.515 & 0.212 & 0.300 & 0.265 & 0.317 \\

{PatchTST\citeyearpar{PatchTST}} 
& 0.387 & 0.400 & 0.281 & 0.326 & 0.469 & 0.455 & 0.387 & 0.407 & 0.216 & 0.304 & 0.259 & 0.281 \\

{iTransformer\citeyearpar{iTransformer}} 
& 0.407 & 0.410 & 0.288 & 0.332 & 0.454 & 0.448 & 0.383 & 0.407 & 0.178 & 0.270 & 0.258 & 0.278 \\

{TimeMixer\citeyearpar{wang2024timemixer}} 
& 0.381 & 0.396 & 0.275 & 0.323 & 0.447 & 0.440 & 0.365 & 0.395 & 0.182 & 0.273 & 0.240 & 0.272 \\

{SimpleTM \citeyearpar{chen2025simpletm}} 
& 0.381 & 0.396 & 0.275 & 0.322 & 0.423 & 0.428 & 0.353 & 0.391 & 0.166 & 0.261 & 0.243 & 0.271 \\

\midrule

\rowcolor{tabhighlight} {Moirai-small} 
& 0.448 & 0.410 & 0.300 & 0.341 & 0.416 & 0.427 & 0.355 & 0.381 & 0.233 & 0.320 & 0.268 & 0.279 \\

\hspace{0.5em} \textit{+ Full finetuning}
& 0.367 & 0.382 & 0.273 & 0.316 & 0.415 & 0.428 & 0.352 & 0.378 & 0.193 & 0.279 & 0.228 & 0.254 \\ 

\hspace{0.5em} \textit{+ Prompt tuning}
& 0.384 & 0.391 & 0.291 & 0.334 & 0.414 & 0.428 & 0.354 & 0.381 & 0.217 & 0.304 & 0.235 & 0.258 \\

\hspace{0.5em} \textit{+ LoRA}
& 0.370 & 0.383 & 0.272 & 0.314 & 0.414 & 0.427 & 0.354 & 0.380 & 0.192 & 0.279 & 0.224 & 0.252 \\

\hspace{0.5em} \textit{+ MSFT \citeyearpar{MSFT}}
& 0.353 & 0.377 & 0.250 & 0.301 & 0.412 & 0.426 & 0.349 & 0.375 & 0.187 & 0.275 & 0.215 & 0.248 \\

\midrule

\rowcolor{tabhighlight} {Moirai-base} 
& 0.382 & 0.388 & 0.281 & 0.326 & 0.412 & 0.424 & 0.356 & 0.388 & 0.188 & 0.274 & 0.246 & 0.265 \\

\hspace{0.5em} \textit{+ Full finetuning}
& 0.368 & 0.371 & 0.258 & 0.307 & 0.408 & 0.424 & 0.357 & 0.385 & 0.173 & 0.264 & 0.231 & 0.258 \\ 

\hspace{0.5em} \textit{+ Prompt tuning}
& 0.378 & 0.386 & 0.279 & 0.325 & 0.411 & 0.423 & 0.360 & 0.387 & 0.183 & 0.271 & 0.230 & 0.255 \\

\hspace{0.5em} \textit{+ LoRA}
& 0.361 & 0.370 & 0.259 & 0.307 & 0.408 & 0.423 & 0.356 & 0.387 & 0.172 & 0.263 & 0.230 & 0.260 \\

\hspace{0.5em} \textit{+ MSFT \citeyearpar{MSFT}}
& 0.332 & 0.369 & 0.247 & 0.305 & 0.407 & 0.422 & 0.352 & 0.383 & 0.169 & 0.260 & 0.213 & 0.244 \\

\midrule
\rowcolor{lightgreen} DGF with Moirai-small 
& 0.338 & 0.369 & 0.238 & 0.292 & 0.399 & 0.420 & \textbf{0.337} & \textbf{0.368} & 0.177 & 0.265 & 0.207 & 0.235 \\

\rowcolor{lightgreen} DGF with Moirai-base 
& \textbf{0.317} & \textbf{0.357} & \textbf{0.234} & \textbf{0.290} & \textbf{0.395} & \textbf{0.410} & 0.340 & 0.374 & \textbf{0.157} & \textbf{0.246} & \textbf{0.203} & \textbf{0.232} \\

\bottomrule
\end{tabular}%
}
\end{table*}

\paragraph{Probabilistic Forecasting}
We further evaluate distributional forecast quality using CRPS and MSIS, which measure the accuracy and calibration of the predictive distribution. As shown in Table~\ref{tab:pf_summary}, DGF achieves the best or competitive results across most datasets, suggesting that mode-conditioned sampling improves not only individual forecasts but also the quality of the generated predictive distribution. Additional probabilistic forecasting metrics are reported in Table~\ref{tab:pf_full}.

\begin{table*}[ht]
  \centering
  \caption{
Probabilistic forecasting results measured by CRPS and MSIS. Lower values are better, and the best results are highlighted in \textbf{bold}.
}
  
  \label{tab:pf_summary}%
\resizebox{0.95\textwidth}{!}{
\begin{tabular}{lccccccccccccc}

\toprule
\multirow{2}[2]{*}{\centering Method} &  \multicolumn{2}{c}{\textbf{Electricity}} & \multicolumn{2}{c}{\textbf{Solar}} &  \multicolumn{2}{c}{\textbf{Weather}} & \multicolumn{2}{c}{\textbf{Istanbul Traffic}} & \multicolumn{2}{c}{\textbf{Turkey Power}} \\
\cmidrule(lr){2-3}  \cmidrule(lr){4-5}  \cmidrule(lr){6-7}  \cmidrule(lr){8-9}  \cmidrule(lr){10-11}  
& CRPS & MSIS & CRPS & MSIS & CRPS & MSIS & CRPS & MSIS & CRPS & MSIS  \\
\midrule

{DeepAR\citeyearpar{salinas2020deepar}} & 0.065 & 6.893 & 0.431 & 11.181  & 0.132 & 21.651 & 0.108 & 4.094 & 0.066 & 13.520 \\

{PatchTST\citeyearpar{PatchTST}} & 0.052 & 5.744 & 0.518 & 8.447 & 0.059 & 7.759 & 0.112 & 3.813 & 0.054 & 8.978 \\

{TiDE\citeyearpar{TiDE}} & 0.048 & 5.672 & 0.420 & 13.754 & 0.054 & 8.095 & 0.110 & 4.752 & 0.046 & 8.579 \\

\midrule

\rowcolor{tabhighlight} {Moirai-small} & 0.072 & 7.999 & 0.471 & 8.425  & 0.049 & 5.236 & 0.173 & 5.937 & 0.048 & 7.127 \\

\hspace{0.5em} \textit{+ Full finetuning}
&  0.055 & 6.009 & 0.395 & 6.947 & 0.039 & 4.477 & 0.151 & 6.735 & 0.040 & 6.887\\

\hspace{0.5em} \textit{+ Prompt tuning}
& 0.066 & 6.595 & 0.421 & 6.936 & 0.050 & 4.901 & 0.154 & 4.733 & 0.045 & 7.042 \\

\hspace{0.5em} \textit{+ LoRA}
& 0.064 & 6.753  & 0.372 & 6.582 & 0.039 & 4.386 & 0.154 & 4.753 & 0.042 & 7.051 \\

\hspace{0.5em} \textit{+ MSFT \citeyearpar{MSFT}}
& 0.047 & 5.327 & 0.353 & 7.706 & 0.036 & 4.178 & 0.141 & 4.447 & 0.038 & 6.810\\

\midrule 

\rowcolor{tabhighlight} {Moirai-base}  & 0.055 & 6.172 & 0.419 & 7.011  & 0.041 & 5.136 & 0.116 & 4.461 & 0.040 & 6.766   \\

\hspace{0.5em} \textit{+ Full finetuning}
& 0.049 & 5.414 & 0.188 & 4.292 & 0.038 & 5.282 & 0.120 & 7.272 & 0.036 & 6.712\\

\hspace{0.5em} \textit{+ Prompt tuning}
& 0.054 & 6.024 & 0.412 & 6.885 & 0.040 & 5.274 &  0.105 & 3.987 & 0.040 & 6.698 \\

\hspace{0.5em} \textit{+ LoRA}
&  0.051 & 5.651 & 0.382 & 6.745 & 0.037 & 4.904 & 0.113 & 4.752 & 0.036 & 6.744 \\

\hspace{0.5em} \textit{+ MSFT \citeyearpar{MSFT}}
& 0.046 & 5.199 & \textbf{0.142} & 3.464 & 0.035 & 4.603 & 0.098 & \textbf{3.685} & 0.034 & 6.419 \\

\midrule 

\rowcolor{lightgreen} DGF with Moirai-small 
& 0.039 & 5.298 & 0.348 & 6.632 & 0.033 & \textbf{4.057} & 0.128 & 4.389 & 0.031 & 6.759 \\

\rowcolor{lightgreen} DGF with Moirai-base
& \textbf{0.031} & \textbf{5.168} & 0.164 & \textbf{3.219} & \textbf{0.030} & 4.159 & \textbf{0.072} & 3.702 & \textbf{0.028} & \textbf{6.264} \\

\bottomrule

\end{tabular}%
}
\end{table*}%

\subsection{Over-Smoothing Analysis}
To examine dynamic-structure preservation, we evaluate generated forecast sets using the STRIPE protocol ~\citep{STRIPE}. We report mean-sample and best-sample MSE and DILATE. Mean-sample scores measure average forecast quality, while best-sample scores measure how closely the best available candidate matches the observed future under a fixed sampling budget. DILATE additionally evaluates shape discrepancy and temporal distortion. As shown in Table~\ref{tab:oversmoothing_stripe}, DGF improves both mean-sample and best-sample scores over the Moirai-base backbone. The mean-sample improvements indicate better overall sample quality, while the best-sample improvements indicate that the generated set contains candidates closer to the observed future. Lower DILATE scores provide evidence of improved dynamic fidelity.

\begin{table*}[t]
\centering
\caption{
Forecasting results on the Traffic and Electricity datasets, averaged over 5 runs (mean $\pm$ std).
Metrics are scaled for readability. Lower values are better.
}
\label{tab:oversmoothing_stripe}
\resizebox{0.98\textwidth}{!}{
\begin{tabular}{lcccccccc}
\toprule
\multirow{2}{*}{Method}
& \multicolumn{4}{c}{Traffic}
& \multicolumn{4}{c}{Electricity} \\
\cmidrule(lr){2-5}
\cmidrule(lr){6-9}
& \multicolumn{2}{c}{MSE}
& \multicolumn{2}{c}{DILATE}
& \multicolumn{2}{c}{MSE}
& \multicolumn{2}{c}{DILATE} \\
\cmidrule(lr){2-3}
\cmidrule(lr){4-5}
\cmidrule(lr){6-7}
\cmidrule(lr){8-9}
& mean & best & mean & best
& mean & best & mean & best \\
\midrule
N-BEATS MSE
& -- & $7.8 \pm 0.3$
& -- & $22.1 \pm 0.8$
& -- & $24.6 \pm 0.9$
& -- & $29.3 \pm 1.3$ \\

N-BEATS DILATE
& -- & $17.1 \pm 0.8$
& -- & $17.8 \pm 0.3$
& -- & $38.9 \pm 1.9$
& -- & $20.7 \pm 0.5$ \\

DeepAR
& $15.1 \pm 1.7$ & $6.6 \pm 0.7$
& $30.3 \pm 1.9$ & $16.9 \pm 0.6$
& $67.6 \pm 5.1$ & $25.6 \pm 0.4$
& $59.8 \pm 5.2$ & $17.2 \pm 0.3$ \\

cVAE-DILATE
& $10.0 \pm 1.7$ & $8.8 \pm 1.6$
& $19.1 \pm 1.2$ & $17.0 \pm 1.1$
& $28.9 \pm 0.8$ & $27.8 \pm 0.8$
& $24.6 \pm 1.4$ & $22.4 \pm 1.3$ \\

Variety Loss
& $9.8 \pm 0.8$ & $7.9 \pm 0.8$
& $18.9 \pm 1.4$ & $15.9 \pm 1.2$
& $29.4 \pm 1.0$ & $27.7 \pm 1.0$
& $24.7 \pm 1.1$ & $21.6 \pm 1.0$ \\

Entropy Regularization
& $11.4 \pm 1.3$ & $10.3 \pm 1.4$
& $19.1 \pm 1.4$ & $16.8 \pm 1.3$
& $34.4 \pm 4.1$ & $32.9 \pm 3.8$
& $29.8 \pm 3.6$ & $25.6 \pm 3.1$ \\

Diverse DPP
& $11.2 \pm 1.8$ & $6.9 \pm 1.0$
& $20.5 \pm 1.0$ & $14.7 \pm 1.0$
& $31.5 \pm 0.8$ & $25.8 \pm 1.3$
& $26.6 \pm 1.0$ & $19.4 \pm 1.0$ \\

STRIPE S+T
& $10.1 \pm 0.4$ & $6.5 \pm 0.2$
& $19.2 \pm 0.8$ & $14.2 \pm 0.2$
& $29.7 \pm 0.3$ & $23.4 \pm 0.2$
& $24.4 \pm 0.3$ & $16.9 \pm 0.2$ \\

\midrule
Moirai-base
& $0.420 \pm 0.01$ & $0.375 \pm 0.01$
& $0.232 \pm 0.01$ & $0.208 \pm 0.01$
& $0.202 \pm 0.01$ & $0.182 \pm 0.01$
& $0.111 \pm 0.01$ & $0.046 \pm 0.01$ \\

Full finetuning
& $0.407 \pm 0.01$ & $0.353 \pm 0.01$
& $0.174 \pm 0.01$ & $0.158 \pm 0.01$
& $0.171 \pm 0.01$ & $0.148 \pm 0.01$
& $0.085 \pm 0.01$ & $0.039 \pm 0.01$ \\

\rowcolor{lightgreen} DGF
& $0.295 \pm 0.01$ & $0.247 \pm 0.01$
& $0.142 \pm 0.01$ & $0.120 \pm 0.01$
& $0.128 \pm 0.01$ & $0.103 \pm 0.01$
& $0.052 \pm 0.01$ & $0.021 \pm 0.01$ \\
\bottomrule
\end{tabular}
}
\end{table*}

\subsection{Prediction Result Visualization}
\begin{figure*}[htpb]
\centering
    \subfloat[ETTm1]{
        \includegraphics[width=0.32\linewidth]{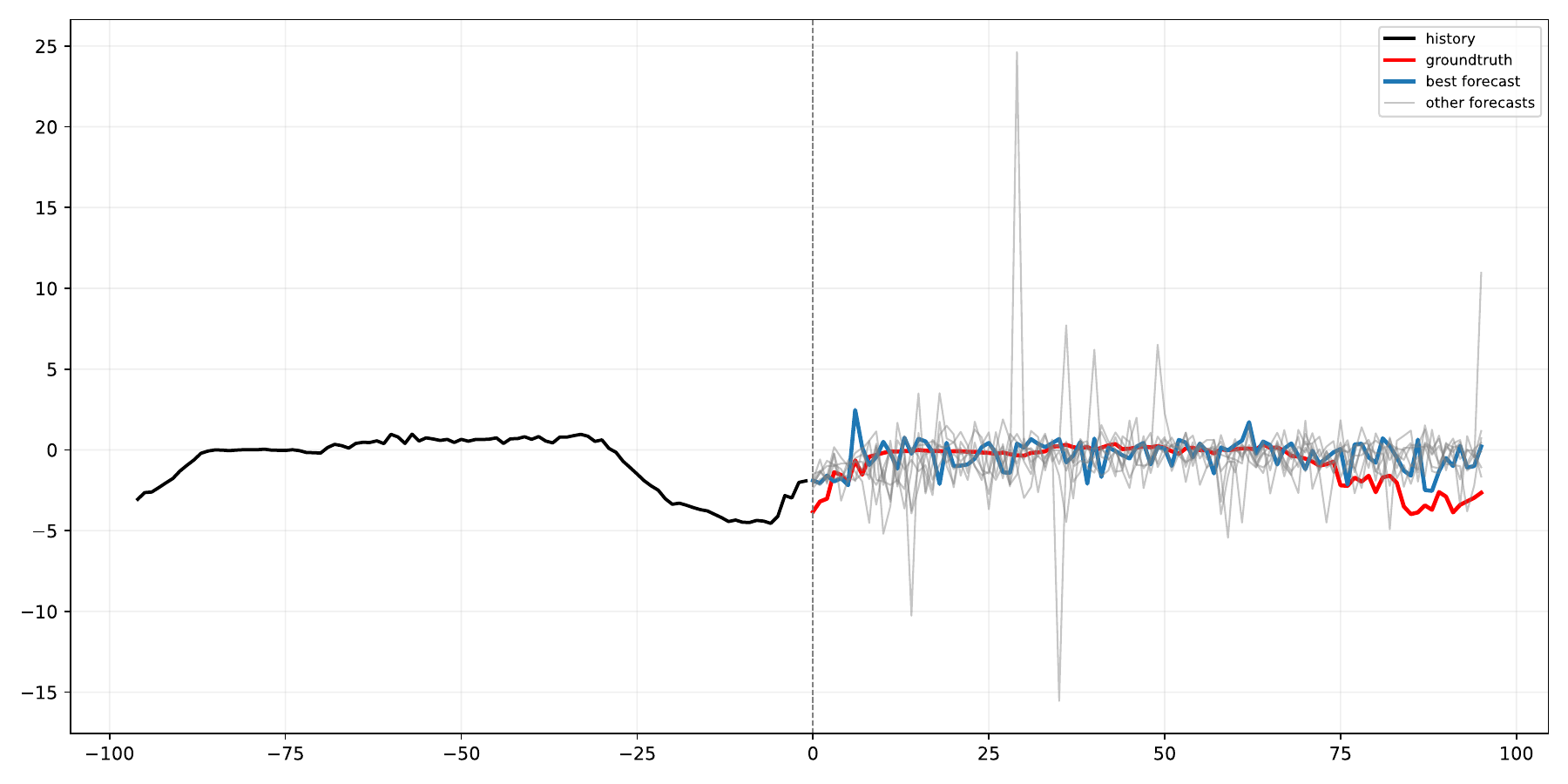}
    }
    \subfloat[ETTm2]{
        \includegraphics[width=0.32\linewidth]{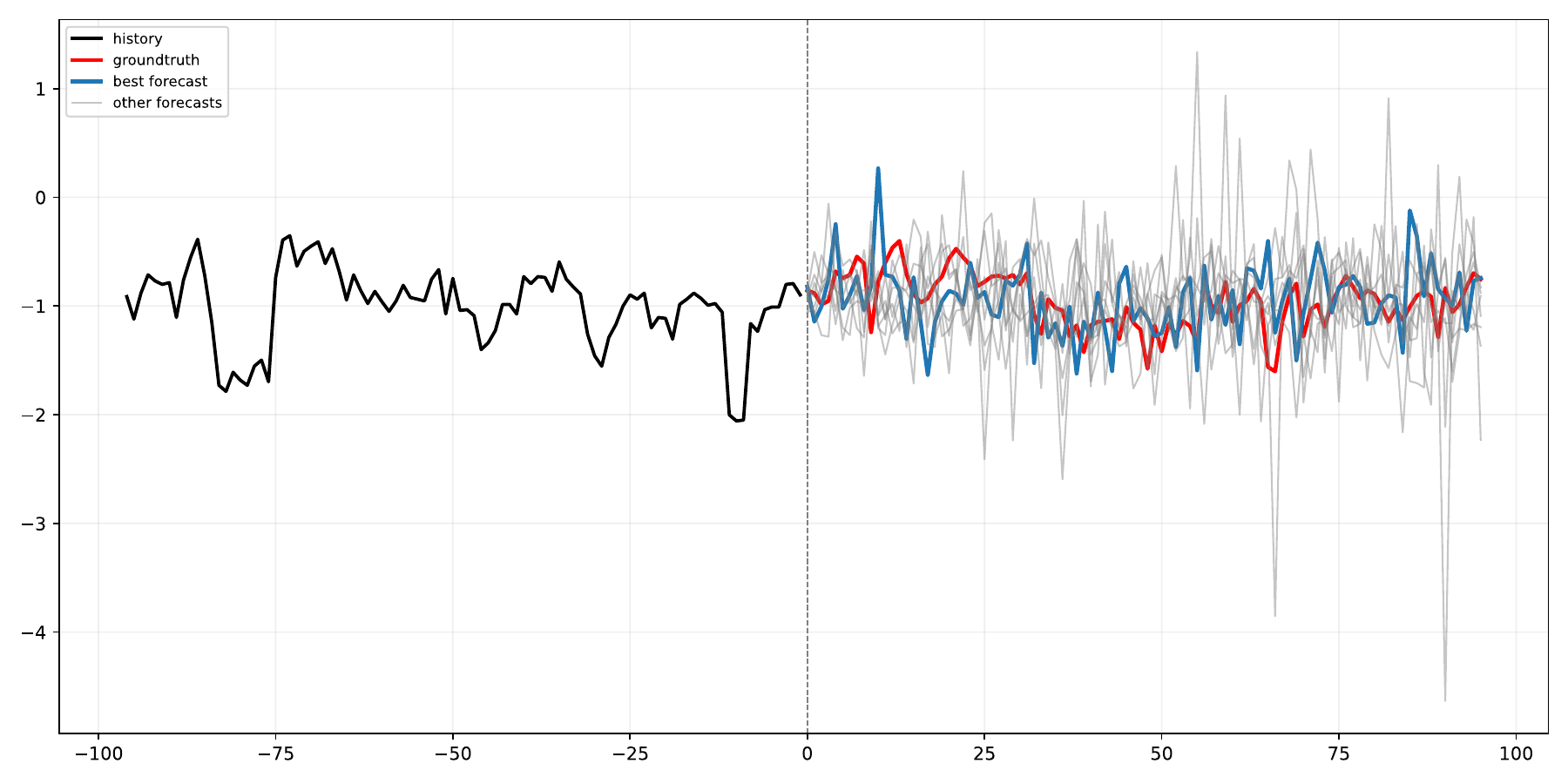}
    }
    \subfloat[ETTh1]{
        \includegraphics[width=0.32\linewidth]{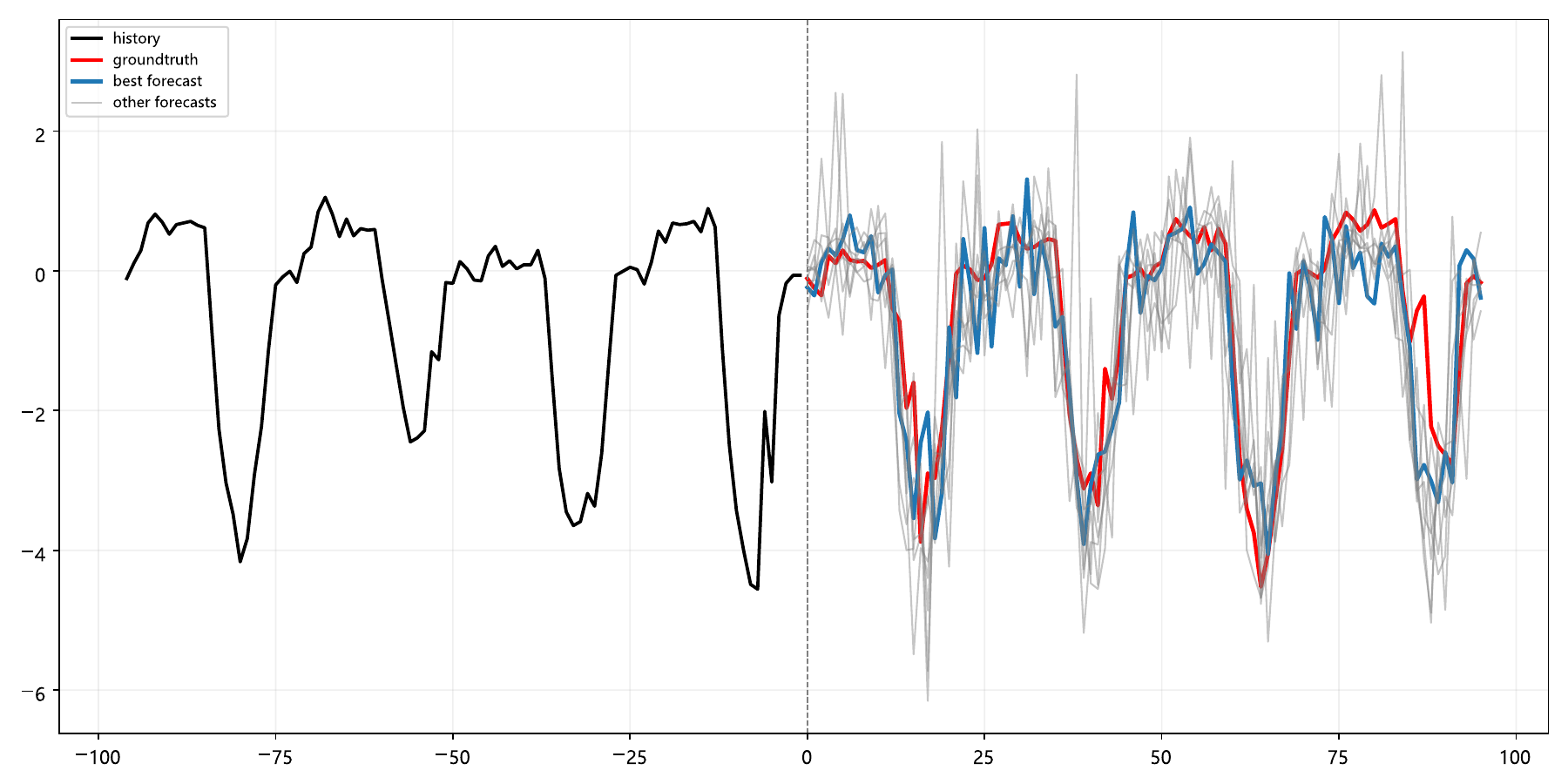}
    }
    \\
    \subfloat[ETTh2]{
        \includegraphics[width=0.32\linewidth]{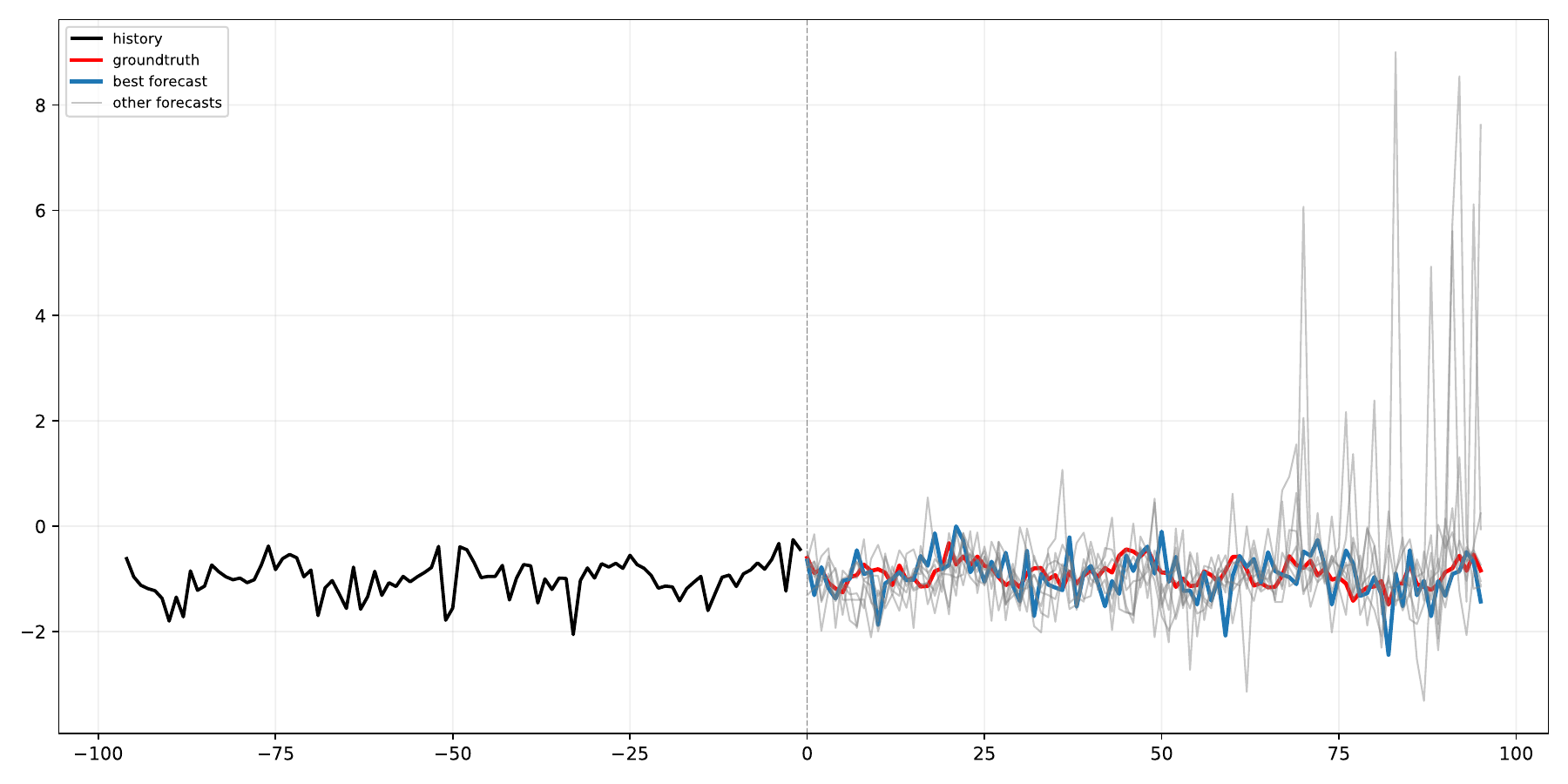}
    }
    \subfloat[Electricity]{
        \includegraphics[width=0.32\linewidth]{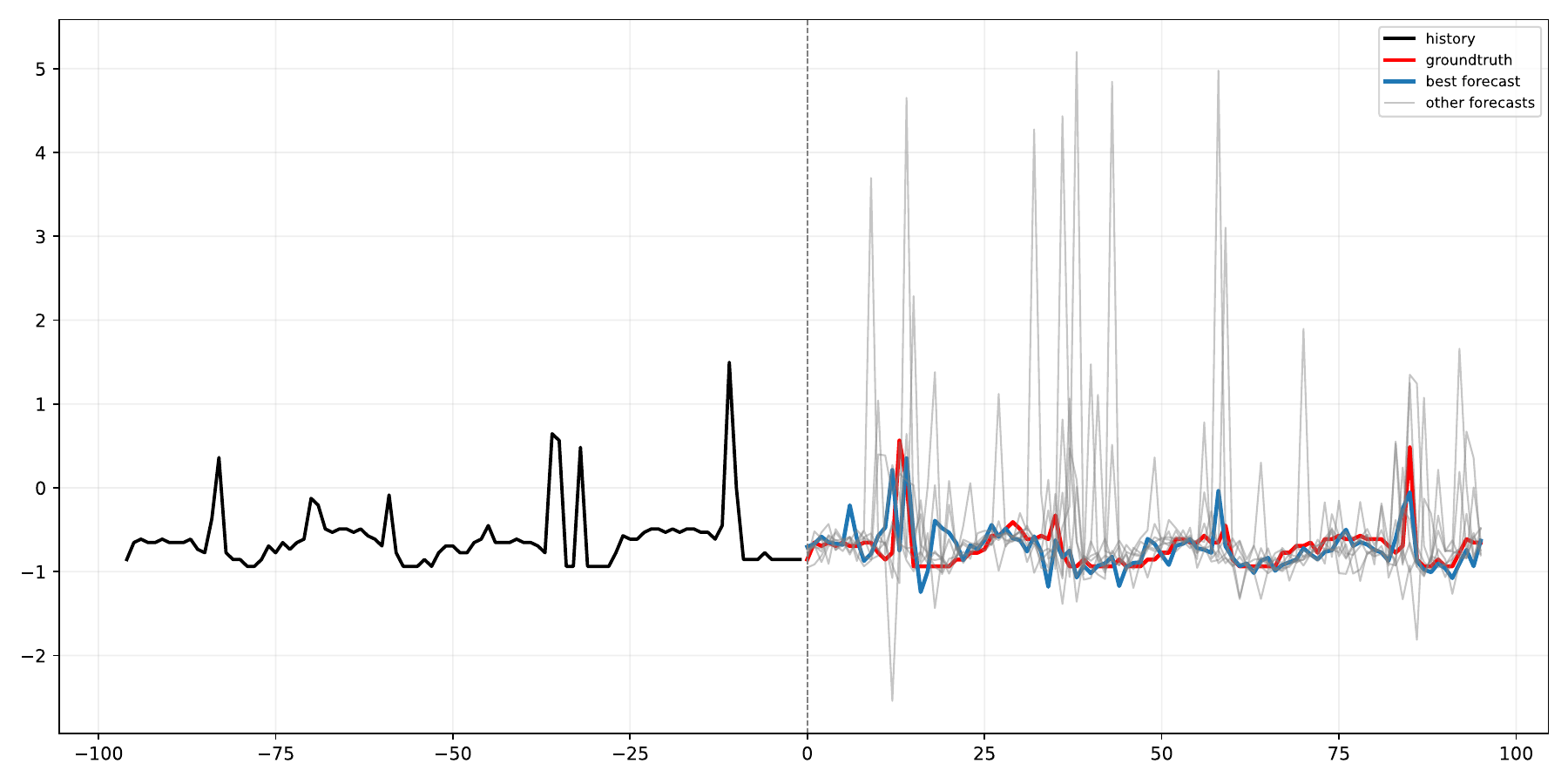}
    }
    \subfloat[Weather]{
        \includegraphics[width=0.32\linewidth]{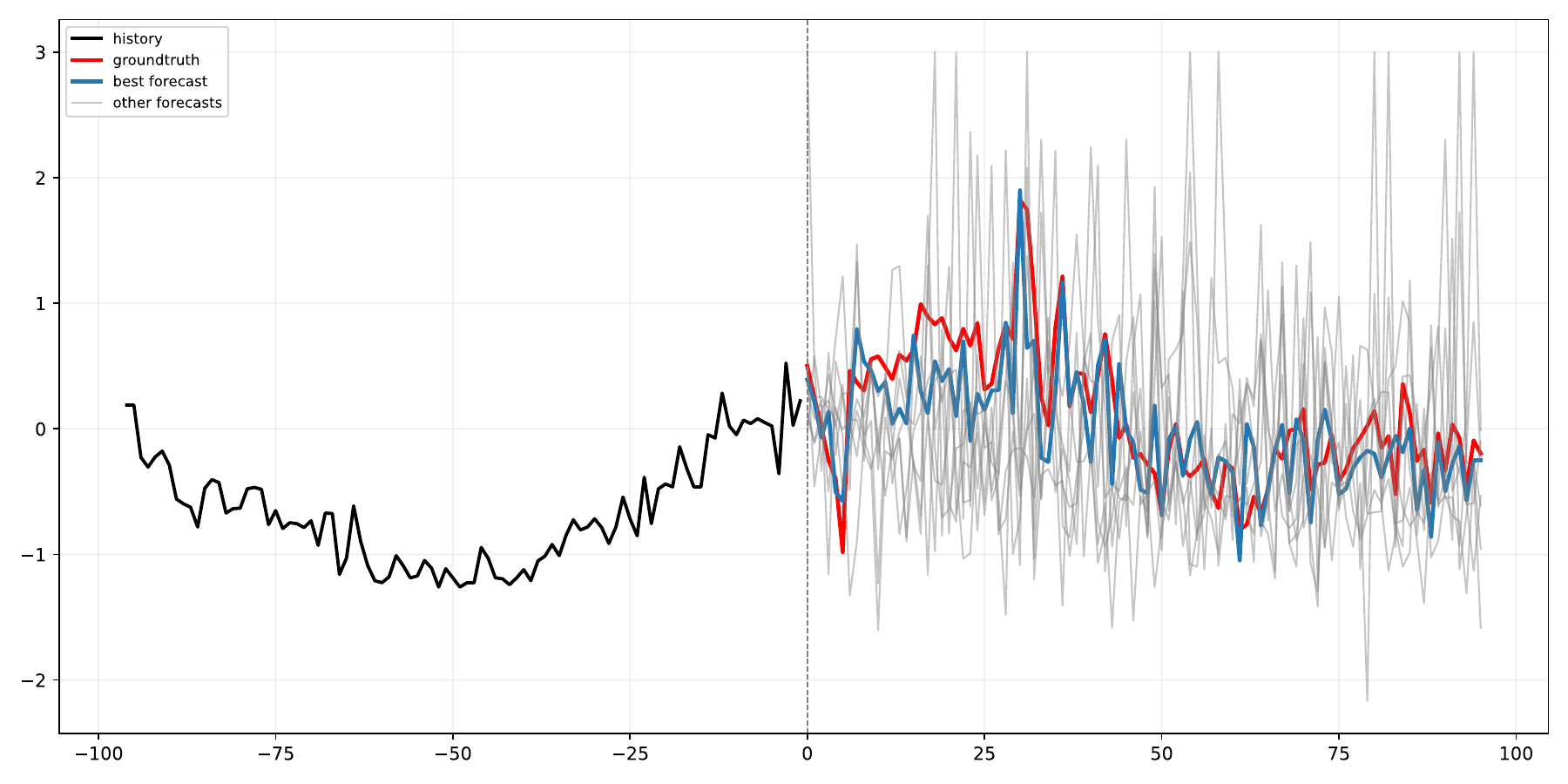}
    }
    \caption{Qualitative visualization of DGF forecasts on six datasets with prediction horizon 96. Each panel shows the last 96 historical observations, the observed future, and eight sampled forecasts, illustrating variation among the generated trajectories.}
\label{fig:case_study}
\end{figure*}

Figure~\ref{fig:case_study} visualizes eight forecast samples for the same input on six datasets, using a prediction horizon of 96. Each panel includes the last 96 historical observations and the observed future. The samples illustrate variation in predicted levels and local temporal patterns. This visualization complements the quantitative sample-quality and dynamic-fidelity evaluation in Section 6.2.

\section{Conclusion}
In this paper, we study over-smoothing in time series forecasting from the perspective of latent dynamical mode compression. We propose DGF, which combines mode-conditioned predictive distributions, Dirichlet-guided group sampling, and hierarchical reward-based optimization with component-diversity regularization. For point forecasting, DGF selects a component mean rather than averaging across components. Experiments on long-sequence and probabilistic forecasting benchmarks demonstrate improved forecasting accuracy and sample quality, with DILATE results providing evidence of better dynamic-structure preservation.

\begin{ack}
The authors sincerely thank the anonymous reviewers for their valuable comments and constructive feedback. This work was supported by the National Natural Science Foundation of China under Grant No. 62506355.
\end{ack}

\bibliography{reference}
\bibliographystyle{plainnat}

\newpage
\appendix

\section{Limitations and Broader Impact}
\label{sec_app:discussion_impact_limitation}

\paragraph{Limitations}
Although DGF is a model-agnostic framework for mode-preserving forecasting, it still has several limitations. First, the reward-based optimization stage benefits from a reasonably good initialization. Since the rewards are defined over sampled trajectories, starting from a poorly trained forecasting model may lead to inefficient exploration and unstable optimization. In this work, we use pretrained time series backbones as a warm start to make the hierarchical sampling process stable and reproducible. This choice is not required by the DGF formulation itself, and the framework can in principle be applied to other forecasting backbones or trained after a supervised pretraining stage. However, the quality of the initialization may affect optimization efficiency and final performance.
Second, DGF introduces additional sampling cost compared with deterministic single-trajectory forecasting. During training, the model samples Dirichlet mode-composition vectors, mode indices, and mode-conditioned trajectories, and hierarchical GRPO requires multiple samples or mini-groups to estimate relative advantages. This increases computation and memory usage. We view this overhead as the cost of preserving multiple latent future modes rather than collapsing them into a single averaged forecast. In practice, this cost can be controlled by the number of modes, samples, and mini-groups, and future work can further reduce it through efficient sampling, mode pruning, or distillation.

\paragraph{Broader Impact.}
This work proposes a mode-preserving forecasting paradigm for time series prediction. Instead of forcing a model to output a single averaged future, DGF encourages the model to represent multiple plausible future modes and reason about their selection uncertainty. This perspective may benefit applications where dynamic events, regime shifts, or rare but important future patterns matter, such as energy management, weather forecasting, traffic control, and financial risk monitoring. By producing forecasts that are more dynamically consistent and mode-diverse, the proposed framework may help downstream decision-makers better understand uncertainty and avoid over-reliance on over-smoothed predictions. At the same time, forecasts generated by DGF should still be interpreted with caution in high-stakes domains. The generated candidates reflect model-based estimates rather than guaranteed future outcomes, and incorrect or overconfident use of multi-modal forecasts may still lead to poor decisions. Appropriate calibration, domain validation, and human oversight remain necessary when deploying such forecasting systems in practice.

\section{Use of Large Language Models}
Large language models (LLMs) were used solely as a general-purpose tool for grammar and language polishing. They were not involved in research ideation, methodology, analysis, or substantive writing. The authors take full responsibility for all contents of the paper.

\section{Notation List}
We summarize the used notation in this paper as
\begin{itemize}
    \item $X=[x_1,\dots,x_T]\in\mathbb{R}^{T}$: historical input time series window with length $T$.

    \item $Y=[x_{T+1},\dots,x_{T+S}]\in\mathbb{R}^{S}$: future trajectory over prediction horizon $S$.

    \item $Y_{\mathrm{gt}}$: observed ground-truth future trajectory paired with the historical input $X$.

    \item $\hat{Y}$: predicted future trajectory generated by the forecasting model.

    \item $\mathcal{D}=\{(X^{(n)},Y^{(n)})\}_{n=1}^{N}$: training dataset containing $N$ historical-future pairs.

    \item $T$: length of the historical input window.

    \item $S$: length of the prediction horizon.

    \item $M\in\{1,\dots,K\}$: discrete latent dynamical mode that represents a possible future regime or dynamic pattern.

    \item $K$: number of latent modes or mode-conditioned predictive distributions.

    \item $p(Y\mid X)$: conditional future distribution given historical input $X$.

    \item $p(Y\mid X,M=m)$: mode-conditioned future distribution under latent mode $m$.

    \item $p(M=m\mid X)$: probability that the future follows the $m$-th latent mode given $X$.

    \item $q_{\theta,k}(Y\mid X)$: the $k$-th mode-conditioned predictive distribution parameterized by model parameters $\theta$.

    \item $h=F_{\mathrm{enc}}(X)$: contextual representation extracted from the historical input by the encoder.

    \item $g_{\theta,k}(\cdot)$: prediction head that maps the contextual representation $h$ to the $k$-th mode-conditioned predictive distribution.

    \item $\boldsymbol{\pi}=[\pi_1,\dots,\pi_K]$: mode-selection probability vector on the $(K-1)$-simplex.

    \item $\Delta^{K-1}$: $(K-1)$-dimensional probability simplex, defined by $\pi_k\geq 0$ and $\sum_{k=1}^{K}\pi_k=1$.

    \item $\boldsymbol{\alpha}=[\alpha_1,\dots,\alpha_K]$: Dirichlet concentration parameter vector.

    \item $\alpha_\theta(X)=[\alpha_{\theta,1}(X),\dots,\alpha_{\theta,K}(X)]$: input-dependent Dirichlet concentration vector predicted by DGF.

    \item $\alpha_0=\sum_{k=1}^{K}\alpha_k$: total concentration of the Dirichlet distribution, controlling dispersion around its mean.

    \item $e_\theta(X)$: evidence score vector used to parameterize the Dirichlet concentration parameters.

    \item $\mathrm{Dir}(\boldsymbol{\pi}\mid\boldsymbol{\alpha})$: Dirichlet distribution over probability vector $\boldsymbol{\pi}$ with concentration parameter $\boldsymbol{\alpha}$.

    \item $\pi_b\sim\mathrm{Dir}(\alpha_\theta(X))$: sampled mode-selection probability vector for the $b$-th mini-group.

    \item $B$: number of mini-groups sampled for the same input.

    \item $G$: number of forecast samples generated within each mini-group.

    \item $z_{b,i}\sim\mathrm{Cat}(\pi_b)$: latent mode index sampled for the $i$-th trajectory in the $b$-th mini-group.

    \item $\hat{Y}_{b,i}$: sampled forecast trajectory generated from the selected mode-conditioned predictive distribution in the $b$-th mini-group.

    \item $p_\theta(\pi_b,z_{b,i},\hat{Y}_{b,i}\mid X)$: joint probability of the hierarchical sampling process for a sampled mode-selection vector, mode index, and forecast trajectory.

    \item $\Phi(\cdot)$: dynamic feature mapping that extracts trajectory-level structural properties such as trend, volatility, frequency pattern, turning points, and transition behavior.

    \item $\Psi(X,\hat{Y})$: transition feature mapping that characterizes the dynamic transition from historical input $X$ to generated future $\hat{Y}$.

    \item $d_{\mathrm{val}}(\hat{Y},Y_{\mathrm{gt}})$: value-level discrepancy between a forecast and the observed future.

    \item $d_{\mathrm{dyn}}(\Phi(\hat{Y}),\Phi(Y_{\mathrm{gt}}))$: dynamic-structure discrepancy between the generated trajectory and the observed future in dynamic feature space.

    \item $d_{\mathrm{trans}}(\Psi(X,\hat{Y}))$: transition discrepancy that penalizes dynamically implausible transitions from history to future.

    \item $\mathcal{Y}\subseteq\mathbb{R}^{S}$: trajectory space over the prediction horizon.

    \item $\mathcal{S}_m\subseteq\mathcal{Y}$: set of dynamically valid trajectories under latent mode $m$.

    \item $\mathcal{S}=\bigcup_{m=1}^{K}\mathcal{S}_m$: union of all dynamically valid trajectory sets.

    \item $r^{\mathrm{traj}}_{b,i}$: trajectory-level reward assigned to the $i$-th sampled forecast in the $b$-th mini-group.

    \item $r_{\mathrm{acc},b,i}$: accuracy reward measuring closeness between $\hat{Y}_{b,i}$ and $Y_{\mathrm{gt}}$.

    \item $r_{\mathrm{dyn},b,i}$: dynamics reward measuring dynamic consistency and transition plausibility of $\hat{Y}_{b,i}$.

    \item $\mathcal{R}_{\mathrm{div}}^{K}(X)$: distribution-level diversity regularizer encouraging the $K$ mode-conditioned predictive distributions to remain distinct.

    \item $D_{\mathrm{mode}}(q_{\theta,k},q_{\theta,l})$: distance between two mode-conditioned predictive distributions.

    \item $D_{\mathrm{val}}(\mu_k,\mu_l)$: value-space distance between mean trajectories of two mode-conditioned predictive distributions.

    \item $D_{\mathrm{dyn}}(\Phi(\mu_k),\Phi(\mu_l))$: dynamic-feature-space distance between two mode-conditioned mean trajectories.

    \item $\mu_k(X)=\mathbb{E}_{q_{\theta,k}}[Y]$: mean trajectory of the $k$-th mode-conditioned predictive distribution.

    \item $m$: margin used in the distribution-level diversity regularizer to prevent rewarding arbitrary divergence.

    \item $\lambda_1,\lambda_2,\lambda_3$: weights for accuracy reward, dynamics reward, and distribution-level diversity regularization.

    \item $\eta_1,\eta_2,\eta_3,\eta_4$: weights controlling the two components of the dynamics reward.

    \item $\rho$: coefficient balancing value-space and dynamic-feature-space distances in $D_{\mathrm{mode}}$.

    \item $\mu_b$: mean trajectory-level reward within the $b$-th mini-group.

    \item $\sigma_b$: standard deviation of trajectory-level rewards within the $b$-th mini-group.

    \item $A^{\mathrm{sample}}_{b,i}$: sample-level advantage for the $i$-th forecast in the $b$-th mini-group.

    \item $R_b$: average trajectory-level reward of the $b$-th mini-group.

    \item $\mu_R$: mean mini-group reward across all $B$ mini-groups.

    \item $\sigma_R$: standard deviation of mini-group rewards across all $B$ mini-groups.

    \item $A^{\mathrm{dir}}_b$: Dirichlet-level advantage for the $b$-th mini-group.

    \item $\mathcal{J}_{\mathrm{sample}}$: sample-level GRPO objective for optimizing sampled modes and selected predictive distributions.

    \item $\mathcal{J}_{\mathrm{dir}}$: Dirichlet-level GRPO objective for optimizing mode-selection probability distribution.

    \item $\mathcal{J}_{\mathrm{DGF}}$: final DGF training objective combining sample-level optimization, Dirichlet-level optimization, and distribution-level diversity regularization.

    \item $\gamma$: weight controlling the strength of the Dirichlet-level objective.

    \item $\epsilon$: small constant used for numerical stability in advantage normalization.

    \item $I(\cdot;\cdot)$: mutual information.

    \item $H(M\mid X)$: conditional entropy of the latent mode given the historical input.

    \item $\ell(\cdot,\cdot)$: generic forecasting loss function.

    \item $\theta$: learnable parameters of DGF.
\end{itemize}

\section{Proofs}
\label{app:proofs}

\subsection{Proof of Theorem~\ref{thm:point_risk_invalid}}
\label{app:proof_point_risk_invalid}

\begin{Proof}
Since the dynamically valid trajectory set $\mathcal{S}$ is non-convex, by definition there exist two trajectories $Y_a,Y_b\in\mathcal{S}$ and a coefficient $\lambda\in(0,1)$ such that
\[
\bar{Y}
=
\lambda Y_a+(1-\lambda)Y_b
\notin \mathcal{S}.
\]

We construct a conditional distribution supported only on these two dynamically valid trajectories:
\[
p(Y\mid X=x)
=
\lambda \delta(Y-Y_a)
+
(1-\lambda)\delta(Y-Y_b),
\]
where $\delta(\cdot)$ denotes the Dirac delta distribution. Since $Y_a,Y_b\in\mathcal{S}$, any trajectory sampled from this conditional distribution belongs to $\mathcal{S}$ almost surely:
\[
Y\sim p(Y\mid X=x)
\Rightarrow
Y\in\mathcal{S}
\quad \text{almost surely}.
\]

Under squared loss, the Bayes-optimal point predictor is
\[
f^\star(x)
=
\arg\min_{f(x)}
\mathbb{E}
\left[
\|f(x)-Y\|_2^2
\mid X=x
\right].
\]
For any candidate prediction $a\in\mathcal{Y}$, the conditional risk is
\[
\mathcal{R}(a)
=
\mathbb{E}
\left[
\|a-Y\|_2^2
\mid X=x
\right].
\]
Expanding the squared norm gives
\[
\mathcal{R}(a)
=
\|a\|_2^2
-
2a^\top \mathbb{E}[Y\mid X=x]
+\mathbb{E}[\|Y\|_2^2\mid X=x].
\]
The last term does not depend on $a$. Taking the gradient with respect to $a$ and setting it to zero yields
\[
2a-2\mathbb{E}[Y\mid X=x]=0,
\]
so the unique minimizer is
\[
f^\star(x)=\mathbb{E}[Y\mid X=x].
\]

For the constructed conditional distribution,
\[
\mathbb{E}[Y\mid X=x]=\lambda Y_a+(1-\lambda)Y_b=\bar{Y}.
\]
By construction, $\bar{Y}\notin\mathcal{S}$. Therefore,
\[
f^\star(x)\notin\mathcal{S}.
\]
Thus, there exists a conditional distribution whose possible futures are all dynamically valid, while the squared-loss optimal point forecast is dynamically invalid.
\end{Proof}

\subsection{Proof of Proposition~\ref{prop:head_collapse}}
\label{app:proof_head_collapse}

\begin{Proof}
For a fixed training sample $(X,Y_{gt})$, the naive multi-head loss is
\[
\mathcal{L}_{\mathrm{naive}}(X,Y_{gt})
=
\sum_{k=1}^{K}
\|\mu_k(X)-Y_{gt}\|_2^2.
\]
Each term is nonnegative and depends only on the corresponding head output $\mu_k(X)$. The loss is minimized if and only if every term is minimized. For squared Euclidean loss, the unique minimizer of the $k$-th term is
\[
\mu_k(X)=Y_{gt}.
\]
Therefore,
\[
\mu_1(X)=\mu_2(X)=\cdots=\mu_K(X)=Y_{gt}.
\]

For the population risk, consider the conditional risk at a fixed input $X=x$:
\[
\mathcal{R}_{\mathrm{naive}}(x)=\sum_{k=1}^{K}\mathbb{E}\left[
\|\mu_k(x)-Y\|_2^2\mid X=x\right].
\]
Again, the objective decomposes over heads. For each $k$, the squared-loss Bayes predictor is
\[\mu_k^\star(x)=\mathbb{E}[Y\mid X=x].
\]
Hence,
\[
\mu_1^\star(x)=\mu_2^\star(x)=\cdots=\mu_K^\star(x)=\mathbb{E}[Y\mid X=x].
\]
This holds for every $x$ where the conditional expectation exists. Therefore, naive multi-head supervision admits a collapsed optimum in which all heads represent the same conditional mean.
\end{Proof}

\section{Implementation Details}
\label{sec:app_imple_details}
This section provides dataset descriptions, finetuning baseline configurations, and metric definitions omitted from the main text.

\subsection{Dataset details} 
For long sequence forecasting (LSF), we conduct experiments on six well-established datasets, including the ETT datasets (ETTh1, ETTh2, ETTm1, ETTm2), Weather, and Electricity. We note that these datasets are not included in the pretraining datasets of the TSFMs we evaluated. The key properties of these LSF datasets are detailed in Table \ref{tab:lsf_datasets}.
Following Moirai \citep{Moirai}, we use 5 out-of-distribution datasets for probabilistic forecasting: Electricity, Solar-Power, Jena Weather, Istanbul Traffic\footnote{https://www.kaggle.com/datasets/leonardo00/istanbul-traffic-index}, and Turkey Power\footnote{https://www.kaggle.com/datasets/dharanikra/electrical-power-demand-in-turkey}. Detailed descriptions of these datasets are provided in Table \ref{tab:pf_datasets}.

\subsection{Implementation Details}
All experiments are implemented in PyTorch~\citep{Pytorch} and trained on NVIDIA H100 80GB GPUs. DGF is built on pretrained Moirai-small and Moirai-base~\citep{Moirai}, which are used as warm-start time series backbones. The Dirichlet concentration parameters are obtained by $\alpha_k=\alpha_{min}+\mathrm{softplus}(e_k)$ to ensure positivity. We use the Adam optimizer~\citep{Adam}, with learning rate selected from $\{10^{-4},5\times10^{-5},10^{-5}\}$ and batch size selected from $\{32,64,128\}$ according to validation performance. Training proceeds for up to 200 epochs, with early stopping applied when validation performance does not improve for 15 epochs. The main DGF hyperparameters include the number of mode-conditioned distributions $K$, the number of Dirichlet mini-groups $B$, the number of samples per mini-group $G$, the Dirichlet-level objective weight $\gamma$, the dynamical consistency weight $\lambda_2$, and the distribution-level diversity weight $\lambda_3$. These values are selected on the validation set from $K\in\{4,8,16,32\}$, $B\in\{2,4,8,16\}$, $G\in\{4,8,16,32\}$, $\gamma\in\{0,0.1,0.3,1,3\}$, $\lambda_2\in(0,0.1)$, and $\lambda_3\in(0,0.01)$. We fix $\lambda_1=1$ for the accuracy reward. Inputs are normalized following the Moirai preprocessing protocol, and predictions are rescaled back to the original scale before evaluation. For deterministic LSF evaluation, DGF reports the MAP-mode output by default; other decision protocols are studied in Appendix~\ref{app_ablation}. Each experiment is repeated 5 times with different random seeds.

\subsection{Finetuning baselines}
Full finetuning involves updating all parameters of the pretrained model. We use a small learning rate for stable adaptation.

LoRA \citep{hu2021lora} introduces trainable rank-decomposition matrices into the attention layers, enabling parameter-efficient finetuning by injecting updates into a low-rank subspace. For Moirai, we use the PEFT library to implement LoRA. We apply LoRA to the query, key, and value projection layers. In addition to the LoRA modules, we also allow the output prediction head to be trainable. The LoRA configuration follows standard settings with rank $r=16$ and scaling factor $\alpha=32$.

For Moirai, we implement prompt fine-tuning by introducing trainable soft prompt embeddings, which are prepended to the input tokens in the embedding space. We avoid inserting them into the patch token space, as doing so can interfere with the statistical computation of RevIN and offers less expressive capacity compared to the high-dimensional embedding space. During inference, we discard the prompt embeddings from the encoder output and use only the time series embeddings as final representation for prediction. Similarly, only the output head and the prompt embeddings are finetuned, while all other parameters remain frozen. Prompt length is set to 2 by default.

\subsection{Metric details}
For long sequence forecasting, we follow the standard protocols to use mean square error (MSE) and mean absolute error for evaluation. For probabilistic forecasting, we include Continuous Ranked Probability Score (CRPS), Mean Scaled Interval Score (MSIS), absolute percentage error (MAPE), symmetric mean absolute percentage error (sMAPE), mean absolute scaled error (MASE), normalized deviation (ND), and normalized root mean squared error (NRMSE) as metrics. The main text reports CRPS and MSIS, while additional probabilistic metrics are reported in Appendix Table~\ref{tab:pf_full}. The definitions and calculations of probabilistic forecasting metrics are as follows. Note that the notations used here are independent of those in the main text.

\paragraph{Continuous Ranked Probability Score}
Given a predicted distribution with c.d.f. \(F\) and ground truth \(\mathbf{Y}\), the CRPS is defined as: 
\begin{align*}
    \textrm{CRPS} & = \int_0^1 2 \Lambda_{\alpha}(F^{-1}(\alpha), \mathbf{Y}) d\alpha \\
    \Lambda_{\alpha}(q, \mathbf{Y}) & = (\alpha - \textbf{1}_\mathrm{\mathbf{Y} < q})(\mathbf{Y} - q),
\end{align*}
where \(\Lambda_{\alpha}\) is the \(\alpha\)-quantile loss, also known as the pinball loss at quantile level \(\alpha\). In practice, the CRPS is intractable or computationally expensive to compute, and we also want to compute a normalized metric, thus we compute a normalized discrete approximation, the mean weighted sum quantile loss, defined as the average of \(K\) quantiles:
\begin{align*}
    \textrm{CRPS} & \approx \frac{1}{K} \sum_{k=1}^K \textrm{wQL}[\alpha_k] \\
    \textrm{wQL}[\alpha] & = 2 \frac{\sum_{i} \Lambda_{\alpha}(\hat{q}_{i}(\alpha), \mathbf{Y}_i)}{\sum_{i} |\mathbf{Y}_{i}|},
\end{align*}
where $\mathbf{Y}_i$ is the ground truth at at time step \(i\) and \(\hat{q}_t(\alpha)\) is the predicted \(\alpha\)-quantile at time step \(i\). We take \(K = 9, \alpha_1 = 0.1, \alpha_2 = 0.2, \ldots, \alpha_9 = 0.9\) in practice.

\paragraph{Mean Scaled Interval Score}
The MSIS is a metric to evaluate uncertainty around point forecasts. Given an upper bound prediction, \(U_i\), and lower bound prediction \(L_i\), the MSIS is defined as:
\begin{align*}
\textrm{MSIS} &= \frac{1}{H \cdot \left( \frac{1}{n-m}\sum_{i=m+1}^n |\mathbf{Y}_i - \mathbf{Y}_{i-m}| \right)} 
\cdot 
\left[
\sum_{i=1}^H (U_i - L_i)  \right. \\
&\quad +\left. \frac{2}{a}(L_i - \mathbf{Y}_i) \mathbb{1}_{\{\mathbf{Y}_i < L_i\}} + \frac{2}{a}(\mathbf{Y}_i - U_i) \mathbb{1}_{\{\mathbf{Y}_i > U_i\}}
\right]
\end{align*}
where \(a = 0.05\) is the significance level for a 95\% prediction interval, over a forecast horizon of length \(H\), and \(m\) is the seasonal factor.

\paragraph{symmetric Mean Absolute Percentage Error }
The sMAPE is an accuracy measure based on percentage errors, treating over- and under-predictions symmetrically, commonly used in forecasting.
\begin{align*}
    \text{SMAPE} &= \frac{200}{H} \sum_{i=1}^H \frac{|\mathbf{Y}_{i} - \widehat{\mathbf{Y}}_{i}|}{|\mathbf{Y}_{i}| + |\widehat{\mathbf{Y}}_{i}|},
\end{align*}

\paragraph{Mean Absolute Scaled Error}
The MASE is a metric for forecasting accuracy, scaling errors by the in-sample mean absolute error of a naive forecast, ensuring interpretability and comparability.
\begin{align*}
    \text{MASE} &= \frac{1}{H} \sum_{i=1}^H \frac{|\mathbf{Y}_{i} - \widehat{\mathbf{Y}}_{i}|}{\frac{1}{H-s} \sum_{j=s+1}^{H} |\mathbf{Y}_j - \mathbf{Y}_{j-s}|},
\end{align*}

where $s$ is the periodicity of the data. $\mathbf{Y},\widehat{\mathbf{Y}}\in\mathbb{R}^{H\times D}$ are the ground truth and prediction results of the future with $H$ time points and $D$ dimensions. $\mathbf{Y}_{i}$ means the $i$-th future time point.
\paragraph{Normalized Deviation}
The ND measures prediction accuracy by standardizing deviations between predicted and actual values, aiding model evaluation and optimization.
\begin{align*}
ND = \frac{\sum_i |Y_i-\hat{Y}_i|}{\sum_i |Y_i|},
\end{align*}
\paragraph{Normalized Root Mean Squared Error}
The NRMSE quantifies prediction error, enables model comparison, aids optimization, and provides interpretable results in time series forecasting.
\begin{align*}
\mathrm{NRMSE}=\frac{\sqrt{\frac{1}{H} \sum_{i=1}^{H}\left(\mathbf{Y}_{i}-\mathbf{\hat{Y}}_{i}\right)^{2}}}{\max (\mathbf{Y})-\min (\mathbf{Y})}.
\end{align*}

\begin{table*}[htbp]
  \caption{Summary of datasets used in the long sequence forecasting evaluation.}\label{tab:lsf_datasets}
  \centering
  \resizebox{0.85\textwidth}{!}{
  \begin{footnotesize}
  \renewcommand{\multirowsetup}{\centering}
  \begin{tabular}{c|l|ccccc}
    \toprule
    Task & Dataset & Variate & Dataset Size & Predict Length & Frequency & \scalebox{0.8}{Information} \\
    \toprule
    & ETTh1 & 7 & 17420  & \scalebox{0.8}{\{96, 192, 336, 720\}} & Hourly &\scalebox{0.8}{Temperature} \\
    \cmidrule{2-7}
    & ETTh2 & 7  & 17420  & \scalebox{0.8}{\{96, 192, 336, 720\}} & Hourly &\scalebox{0.8}{Temperature} \\
    \cmidrule{2-7}
    Long Sequence & ETTm1 & 7 & 69680 & \scalebox{0.8}{\{96, 192, 336, 720\}} & 15 min &\scalebox{0.8}{Temperature} \\
    \cmidrule{2-7}
    Forecasting &ETTm2 & 7 & 69680 & \scalebox{0.8}{\{96, 192, 336, 720\}} & 15 min &\scalebox{0.8}{Temperature} \\
    \cmidrule{2-7}
     & Electricity & 321 & 26304 & \scalebox{0.8}{\{96, 192, 336, 720\}} & Hourly & \scalebox{0.8}{Electricity} \\
    \cmidrule{2-7}
    & Weather & 21 & 52696 & \scalebox{0.8}{\{96, 192, 336, 720\}} &10 min &\scalebox{0.8}{Weather} \\

    \bottomrule
  \end{tabular}
  \end{footnotesize}
  }
\end{table*}%

\begin{table*}[htbp]
  \caption{Summary of datasets used in the probabilistic forecasting evaluation setting.}\label{tab:pf_datasets}
  \centering
  \resizebox{0.85\textwidth}{!}{
  \begin{footnotesize}
  \renewcommand{\multirowsetup}{\centering}
  \begin{tabular}{c|l|cccccc}
    \toprule
    Task & Dataset & Variate & Dataset Size & Predict Length & Rolling Evaluation &Frequency & \scalebox{0.8}{Information} \\
    \toprule
    & Electricity & 321 & 26304 & 24 & 7 & H &\scalebox{0.8}{Energy} \\
    \cmidrule{2-8}
    Probabilistic & Solar &  137 & 8760 & 24 & 7 & H &\scalebox{0.8}{Energy} \\
    \cmidrule{2-8}
    Forecasting & Weather & 21 & 52696 & 144 & 7 & 10T &\scalebox{0.8}{Climate} \\
    \cmidrule{2-8}
    & Istanbul Traffic & 3 & 14244 & 24 & 7 & H &\scalebox{0.8}{Transport} \\
    \cmidrule{2-8}
    & Turkey Power & 18 & 26304 & 24 & 7 & H &\scalebox{0.8}{Energy} \\
        \bottomrule
  \end{tabular}
  \end{footnotesize}
  }
\end{table*}%

\subsection{Computational Cost Analysis}
\label{app:cost_analysis}

We further analyze the computational cost of DGF to examine whether the proposed mode-preserving forecasting mechanism introduces additional overhead. All cost benchmarks are implemented in PyTorch~\citep{Pytorch} and run with MOIRAI-1.0-small as the backbone. We use context length 96, prediction horizon 24, and \texttt{num\_samples}=8. All results are obtained with seed 1. We compare three settings: zero-shot Moirai, supervised full finetuning, and DGF full finetuning. For both supervised full finetuning and DGF, the trainable ratio is 1.0, i.e., the backbone is not frozen.

The timing protocol is designed to isolate the computational cost of training and inference. The reported training wall-clock time measures only the optimizer-update stage, including forward propagation, loss computation, backward propagation, and optimizer step; model loading, data loading, and initialization are excluded. Inference time excludes model loading and is measured after two warm-up batches. CUDA synchronization is applied before and after timing to reduce measurement noise. Table~\ref{tab:accuracy_cost_tradeoff} shows the accuracy-cost trade-off on ETTh1, ETTm1, and Weather, including training time, inference time, and GPU memory usage. The average MSE values are taken from the main forecasting experiments, whereas the cost measurements use context length 96, prediction horizon 24, and eight samples. Under this configuration, DGF achieves lower average forecasting MSE and comparable or lower measured computation time than supervised full finetuning. These timing results characterize the specified benchmark and do not estimate the total cost of the full experimental campaign.

\begin{table*}[t] 
\centering 
\caption{ Accuracy-cost trade-off with MOIRAI-1.0-small. We report average MSE from the main forecasting results and computational cost from the seed-1 cost benchmark with context length 96, prediction horizon 24, and \texttt{num\_samples}=8. Training time excludes model loading, data loading, and initialization, and measures only optimizer-update computation. Inference time excludes model loading and is measured after two warm-up batches. Lower values are better. }
\label{tab:accuracy_cost_tradeoff} 
\resizebox{\textwidth}{!}{ \begin{tabular}{llccccc} \toprule Dataset & Method & Avg. MSE & Train Total (s) & Infer Total (s) & Train Mem. (GB) & Infer Mem. (GB) \\ \midrule \multirow{3}{*}{ETTh1} & Moirai-small zero-shot & 0.416 & -- & 1.0763 & -- & 0.0852 \\ & Supervised full finetuning & 0.415 & 10.8496 & 1.1620 & 0.3236 & 0.1694 \\ & DGF full finetuning & {0.399} & {9.4667} & {1.0560} & {0.2943} & {0.1620} \\ \midrule \multirow{3}{*}{ETTm1} & Moirai-small zero-shot & 0.448 & -- & 1.0516 & -- & 0.0852 \\ & Supervised full finetuning & 0.367 & 10.5942 & 1.0754 & 0.3236 & 0.1694 \\ & DGF full finetuning & {0.338} & {9.3407} & {1.0467} & {0.2943} & {0.1620} \\ \midrule \multirow{3}{*}{Weather} & Moirai-small zero-shot & 0.268 & -- & 1.0551 & -- & 0.0862 \\ & Supervised full finetuning & 0.228 & 10.6123 & 1.0946 & 0.3237 & 0.1704 \\ & DGF full finetuning & {0.207} & {8.8507} & {1.0629} & {0.2943} & {0.1631} \\ \bottomrule \end{tabular} } 
\end{table*}

\section{Additional Ablation Study} \label{app_ablation}
\paragraph{Component Analysis}
We conduct ablation studies to examine the contribution of each major component in DGF. The variants remove or replace the components corresponding to mode-conditioned representation, Dirichlet-guided mode selection, and reward-based optimization. As shown in Table~\ref{tab:ablation_components}, removing any major component degrades performance. The drops from \textit{w/o multiple heads} and \textit{softmax routing} confirm the importance of explicitly representing multiple latent modes and modeling uncertainty over their selection probabilities. Removing the dynamics reward or diversity regularization weakens performance, indicating that dynamic feedback and mode-level separation are both necessary for mode-preserving forecasting. Finally, removing the Dirichlet-level advantage leads to degradation, showing that group-level credit assignment is important for learning effective mode-composition hypotheses.

\begin{table*}[t]
\centering
\caption{Component ablation of DGF using Moirai-base as the backbone. The ablation is conducted on representative LSF datasets with four prediction lengths.}
\label{tab:ablation_components}
\resizebox{\textwidth}{!}{
\begin{tabular}{@{}lcc>{\columncolor{lightgreen}}ccccccc@{}}
\toprule
Dataset & Len. & Metric & DGF & \shortstack{w/o multiple\\heads} & \shortstack{softmax\\routing} & \shortstack{w/o dynamics\\reward} & \shortstack{w/o diversity\\regularization} & \shortstack{w/o Dirichlet-level\\advantage} & \shortstack{supervised\\adaptation only} \\
\midrule
\multirow{8}{*}{ETTm1} & \multirow{2}{*}{96} & MSE & 0.265 & 0.299 & 0.315 & 0.289 & 0.278 & 0.300 & 0.312 \\
 &  & MAE & 0.322 & 0.330 & 0.336 & 0.326 & 0.326 & 0.330 & 0.334 \\
 & \multirow{2}{*}{192} & MSE & 0.302 & 0.339 & 0.356 & 0.324 & 0.317 & 0.344 & 0.355 \\
 &  & MAE & 0.347 & 0.352 & 0.362 & 0.351 & 0.351 & 0.360 & 0.361 \\
 & \multirow{2}{*}{336} & MSE & 0.330 & 0.370 & 0.379 & 0.356 & 0.342 & 0.373 & 0.380 \\
 &  & MAE & 0.365 & 0.378 & 0.383 & 0.372 & 0.369 & 0.375 & 0.380 \\
 & \multirow{2}{*}{720} & MSE & 0.372 & 0.410 & 0.423 & 0.401 & 0.389 & 0.416 & 0.426 \\
 &  & MAE & 0.392 & 0.406 & 0.408 & 0.400 & 0.399 & 0.405 & 0.409 \\
\midrule
\multirow{8}{*}{ETTm2} & \multirow{2}{*}{96} & MSE & 0.152 & 0.170 & 0.177 & 0.162 & 0.160 & 0.175 & 0.176 \\
 &  & MAE & 0.229 & 0.242 & 0.254 & 0.240 & 0.240 & 0.245 & 0.249 \\
 & \multirow{2}{*}{192} & MSE & 0.202 & 0.224 & 0.228 & 0.215 & 0.211 & 0.225 & 0.230 \\
 &  & MAE & 0.273 & 0.283 & 0.285 & 0.278 & 0.277 & 0.283 & 0.288 \\
 & \multirow{2}{*}{336} & MSE & 0.251 & 0.273 & 0.280 & 0.264 & 0.264 & 0.274 & 0.282 \\
 &  & MAE & 0.308 & 0.319 & 0.328 & 0.320 & 0.312 & 0.322 & 0.325 \\
 & \multirow{2}{*}{720} & MSE & 0.330 & 0.341 & 0.343 & 0.337 & 0.336 & 0.340 & 0.344 \\
 &  & MAE & 0.352 & 0.363 & 0.371 & 0.360 & 0.356 & 0.365 & 0.367 \\
\midrule
\multirow{8}{*}{Electricity} & \multirow{2}{*}{96} & MSE & 0.127 & 0.140 & 0.138 & 0.135 & 0.133 & 0.137 & 0.144 \\
 &  & MAE & 0.219 & 0.230 & 0.239 & 0.226 & 0.223 & 0.234 & 0.236 \\
 & \multirow{2}{*}{192} & MSE & 0.142 & 0.160 & 0.169 & 0.155 & 0.151 & 0.159 & 0.166 \\
 &  & MAE & 0.239 & 0.249 & 0.252 & 0.247 & 0.244 & 0.256 & 0.256 \\
 & \multirow{2}{*}{336} & MSE & 0.165 & 0.175 & 0.176 & 0.169 & 0.169 & 0.173 & 0.176 \\
 &  & MAE & 0.251 & 0.263 & 0.271 & 0.258 & 0.259 & 0.266 & 0.267 \\
 & \multirow{2}{*}{720} & MSE & 0.192 & 0.203 & 0.207 & 0.197 & 0.198 & 0.201 & 0.207 \\
 &  & MAE & 0.274 & 0.290 & 0.298 & 0.284 & 0.280 & 0.292 & 0.295 \\
\bottomrule
\end{tabular}
}
\end{table*}

\paragraph{Forecast decision protocols.}
DGF defines a stochastic forecasting policy, while LSF benchmarks require a single reported trajectory. We compare four protocols: \textit{Random sample}, which evaluates one trajectory sampled from the full DGF process; \textit{MAP-mode output}, which reports the mean trajectory of the mode with the largest expected Dirichlet probability; \textit{Sample-and-rerank}, which selects from sampled candidates using a history-to-future dynamical consistency score without ground-truth access; and \textit{Best-of-N}, an oracle protocol that selects the candidate with the lowest ground-truth error. MAP-mode is used in the main table, while Best-of-N is reported only as a mode-coverage upper bound.

\begin{table*}[t]
\centering
\caption{Ablation on deterministic forecast decision protocols for LSF using Moirai-base as the backbone. MAP-mode output is the reported protocol in the main table. Best-of-N is reported only as an oracle coverage indicator and is not used as the default deterministic protocol.}
\label{tab:ablation_decision_protocol}
\resizebox{\textwidth}{!}{
\begin{tabular}{@{}lcccccc@{}}
\toprule
Dataset & Len. & Metric & \shortstack{MAP-mode output} & \shortstack{Random sample} & \shortstack{Sample-and-rerank} & \shortstack{Best-of-N oracle} \\
\midrule
\multirow{8}{*}{ETTm1} & \multirow{2}{*}{96} & MSE & 0.265 & 0.278 & 0.258 & 0.238 \\
 &  & MAE & 0.322 & 0.331 & 0.317 & 0.303 \\
 & \multirow{2}{*}{192} & MSE & 0.302 & 0.319 & 0.294 & 0.271 \\
 &  & MAE & 0.347 & 0.358 & 0.341 & 0.326 \\
 & \multirow{2}{*}{336} & MSE & 0.330 & 0.347 & 0.321 & 0.297 \\
 &  & MAE & 0.365 & 0.376 & 0.359 & 0.342 \\
 & \multirow{2}{*}{720} & MSE & 0.372 & 0.391 & 0.362 & 0.336 \\
 &  & MAE & 0.392 & 0.407 & 0.385 & 0.368 \\
\midrule
\multirow{8}{*}{ETTm2} & \multirow{2}{*}{96} & MSE & 0.152 & 0.160 & 0.148 & 0.136 \\
 &  & MAE & 0.229 & 0.236 & 0.225 & 0.216 \\
 & \multirow{2}{*}{192} & MSE & 0.202 & 0.214 & 0.197 & 0.181 \\
 &  & MAE & 0.273 & 0.282 & 0.269 & 0.258 \\
 & \multirow{2}{*}{336} & MSE & 0.251 & 0.265 & 0.244 & 0.225 \\
 &  & MAE & 0.308 & 0.318 & 0.303 & 0.291 \\
 & \multirow{2}{*}{720} & MSE & 0.330 & 0.347 & 0.322 & 0.298 \\
 &  & MAE & 0.352 & 0.364 & 0.347 & 0.333 \\
\midrule
\multirow{8}{*}{Electricity} & \multirow{2}{*}{96} & MSE & 0.127 & 0.134 & 0.124 & 0.114 \\
 &  & MAE & 0.219 & 0.226 & 0.216 & 0.207 \\
 & \multirow{2}{*}{192} & MSE & 0.142 & 0.151 & 0.138 & 0.127 \\
 &  & MAE & 0.239 & 0.247 & 0.236 & 0.226 \\
 & \multirow{2}{*}{336} & MSE & 0.165 & 0.175 & 0.160 & 0.147 \\
 &  & MAE & 0.251 & 0.259 & 0.247 & 0.237 \\
 & \multirow{2}{*}{720} & MSE & 0.192 & 0.204 & 0.186 & 0.171 \\
 &  & MAE & 0.274 & 0.284 & 0.269 & 0.257 \\
\bottomrule
\end{tabular}
}
\end{table*}

\subsection{Parameter Sensitivity}
\begin{figure*}[htpb]
\centering
    \subfloat[K]{
        \includegraphics[width=0.23\linewidth]{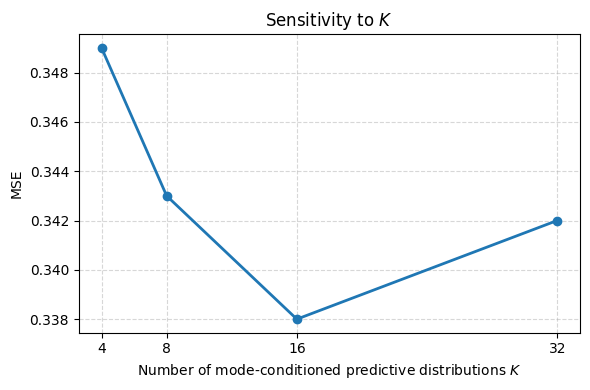}
    }
    \subfloat[B]{
        \includegraphics[width=0.23\linewidth]{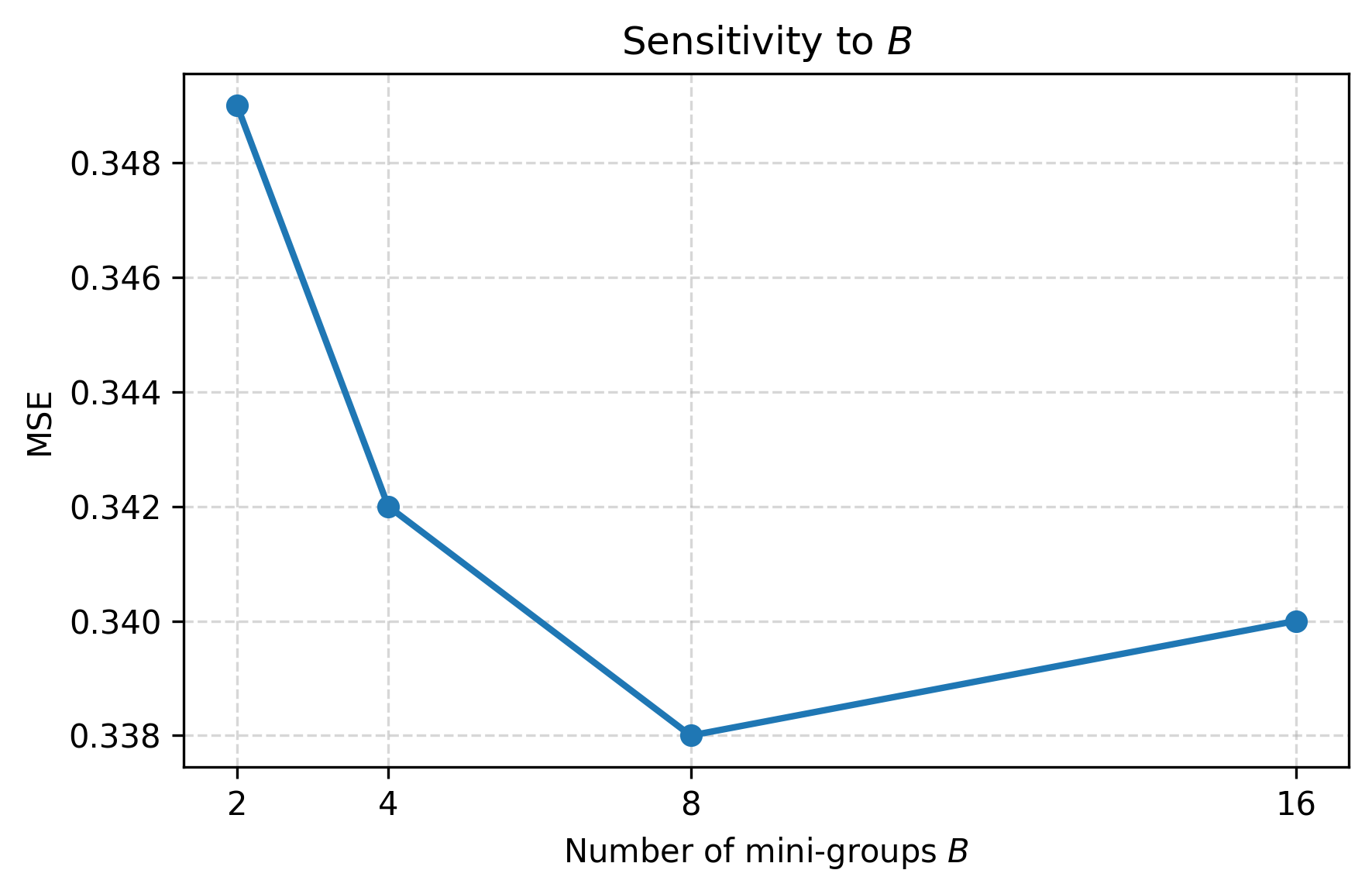}
    }
    \subfloat[G]{
        \includegraphics[width=0.23\linewidth]{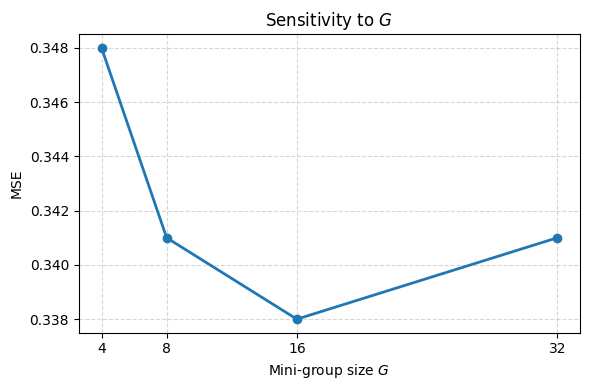}
    }
    \subfloat[$\gamma$]{
        \includegraphics[width=0.23\linewidth]{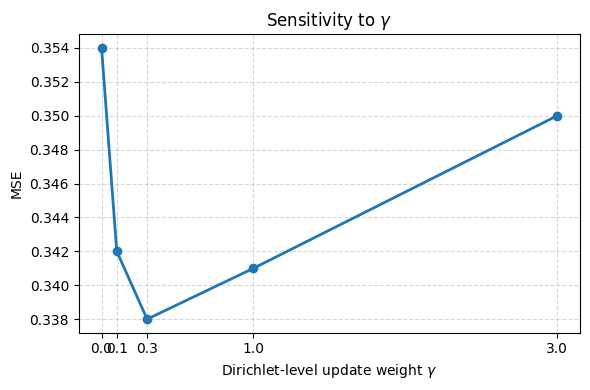}
    }
    \\
    \subfloat[$\alpha_{\min}$]{
        \includegraphics[width=0.23\linewidth]{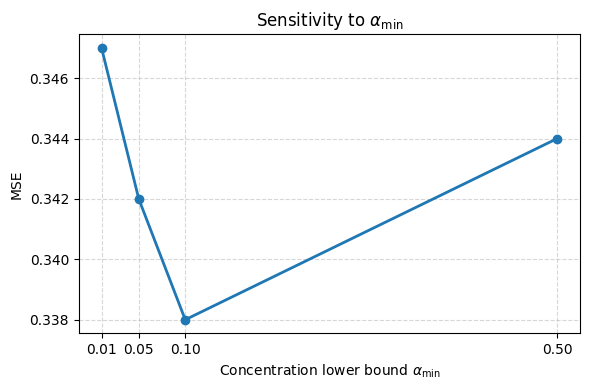}
    }
    \subfloat[$\lambda_2$]{
        \includegraphics[width=0.23\linewidth]{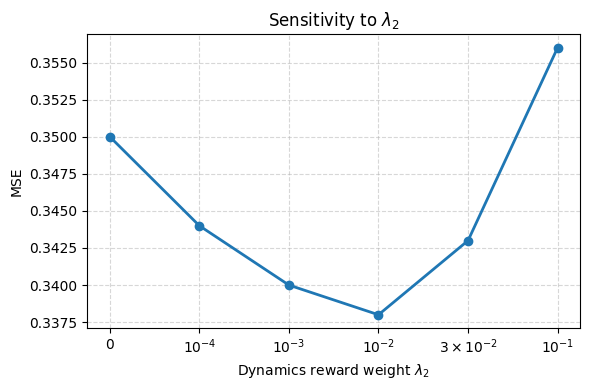}
    }
    \subfloat[$\lambda_3$]{
        \includegraphics[width=0.23\linewidth]{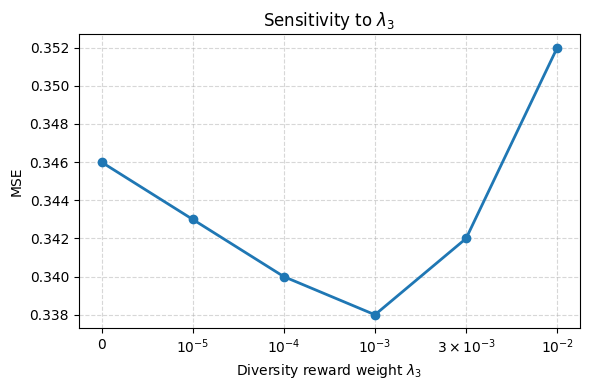}
    }
    \subfloat[Input Length]{
        \includegraphics[width=0.23\linewidth]{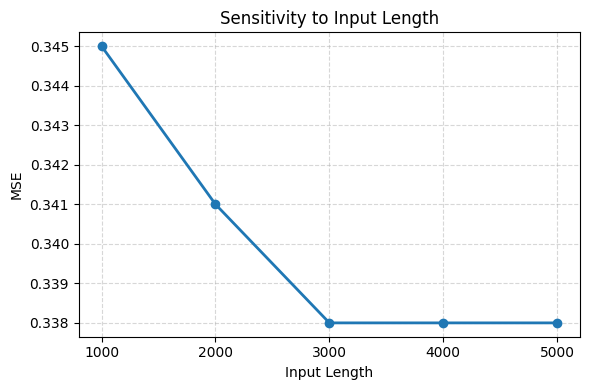}
    }
    
    \caption{Parameter Sensitivity Analysis.}
    \label{fig:psa}
\end{figure*}

We analyze the sensitivity of DGF to major hyperparameters. All experiments use Moirai-small on ETTh1 with prediction length 96. We vary one hyperparameter at a time and report MSE in Figure~\ref{fig:psa}. Overall, DGF remains stable across a reasonable range of hyperparameter values.

\section{Full results}
We provide the full MSE and MAE results in Tables \ref{tab:lsf_full_mse} and \ref{tab:lsf_full_mae}, and the full probabilistic forecasting results in Table \ref{tab:pf_full}.

\begin{table*}[t]
\centering
\caption{Full MSE results of long sequence forecasting experiments.}
\label{tab:lsf_full_mse}
\resizebox{\textwidth}{!}{
\begin{tabular}{@{}lcccccc>{\columncolor{tabhighlight}}ccccc>{\columncolor{tabhighlight}}ccccc>{\columncolor{lightgreen}}c>{\columncolor{lightgreen}}c@{}}
\toprule
 & & \multicolumn{5}{c}{Baselines} & \multicolumn{5}{c}{Moirai-small} & \multicolumn{5}{c}{Moirai-base} & \multicolumn{2}{c}{DGF} \\
\cmidrule(lr){3-7}\cmidrule(lr){8-12}\cmidrule(lr){13-17}\cmidrule(lr){18-19}
Dataset & Len. & \rotatebox{90}{DLinear\,\citeyearpar{DLinear}} & \rotatebox{90}{PatchTST\,\citeyearpar{PatchTST}} & \rotatebox{90}{iTransformer\,\citeyearpar{iTransformer}} & \rotatebox{90}{TimeMixer\,\citeyearpar{wang2024timemixer}} & \rotatebox{90}{SimpleTM\,\citeyearpar{chen2025simpletm}} & \rotatebox{90}{Moirai-small} & \rotatebox{90}{\textit{+ Full finetuning}} & \rotatebox{90}{\textit{+ Prompt tuning}} & \rotatebox{90}{\textit{+ LoRA}} & \rotatebox{90}{\textit{+ MSFT}\,\citeyearpar{MSFT}} & \rotatebox{90}{Moirai-base} & \rotatebox{90}{\textit{+ Full finetuning}} & \rotatebox{90}{\textit{+ Prompt tuning}} & \rotatebox{90}{\textit{+ LoRA}} & \rotatebox{90}{\textit{+ MSFT}\,\citeyearpar{MSFT}} & \rotatebox{90}{DGF with Moirai-small} & \rotatebox{90}{DGF with Moirai-base} \\
\midrule
\multirow{4}{*}{\rotatebox{90}{ETTm1}} & 96 & 0.345 & 0.329 & 0.334 & 0.320 & 0.321 & 0.404 & 0.303 & 0.335 & 0.302 & 0.295 & 0.335 & 0.312 & 0.330 & 0.311 & 0.284 & 0.280 & \textbf{0.265} \\
 & 192 & 0.380 & 0.367 & 0.377 & 0.361 & 0.360 & 0.435 & 0.352 & 0.368 & 0.357 & 0.338 & 0.366 & 0.355 & 0.363 & 0.345 & 0.317 & 0.321 & \textbf{0.302} \\
 & 336 & 0.413 & 0.399 & 0.426 & 0.390 & 0.390 & 0.462 & 0.388 & 0.405 & 0.389 & 0.371 & 0.391 & 0.380 & 0.389 & 0.373 & 0.343 & 0.358 & \textbf{0.330} \\
 & 720 & 0.474 & 0.454 & 0.491 & 0.454 & 0.454 & 0.490 & 0.425 & 0.428 & 0.431 & 0.409 & 0.434 & 0.426 & 0.431 & 0.414 & 0.382 & 0.392 & \textbf{0.372} \\
\midrule
\multirow{4}{*}{\rotatebox{90}{ETTm2}} & 96 & 0.193 & 0.175 & 0.180 & 0.175 & 0.173 & 0.205 & 0.179 & 0.197 & 0.179 & 0.165 & 0.197 & 0.176 & 0.197 & 0.177 & 0.166 & 0.154 & \textbf{0.152} \\
 & 192 & 0.284 & 0.241 & 0.250 & 0.237 & 0.238 & 0.261 & 0.234 & 0.252 & 0.234 & 0.218 & 0.250 & 0.230 & 0.247 & 0.230 & 0.217 & 0.205 & \textbf{0.202} \\
 & 336 & 0.369 & 0.305 & 0.311 & 0.298 & 0.296 & 0.319 & 0.291 & 0.304 & 0.288 & 0.267 & 0.301 & 0.282 & 0.300 & 0.280 & 0.265 & 0.257 & \textbf{0.251} \\
 & 720 & 0.554 & 0.402 & 0.412 & 0.391 & 0.393 & 0.415 & 0.388 & 0.413 & 0.387 & 0.349 & 0.375 & 0.344 & 0.374 & 0.347 & 0.339 & 0.338 & \textbf{0.330} \\
\midrule
\multirow{4}{*}{\rotatebox{90}{ETTh1}} & 96 & 0.386 & 0.414 & 0.386 & 0.375 & 0.366 & 0.387 & 0.382 & 0.384 & 0.382 & 0.380 & 0.375 & 0.372 & 0.375 & 0.373 & 0.372 & 0.364 & \textbf{0.361} \\
 & 192 & 0.437 & 0.460 & 0.441 & 0.429 & 0.422 & 0.418 & 0.419 & 0.415 & 0.418 & 0.416 & 0.406 & 0.404 & 0.406 & 0.404 & 0.404 & 0.401 & \textbf{0.391} \\
 & 336 & 0.481 & 0.501 & 0.487 & 0.484 & 0.440 & 0.431 & 0.434 & 0.429 & 0.431 & 0.428 & 0.426 & 0.423 & 0.425 & 0.423 & 0.422 & 0.419 & \textbf{0.414} \\
 & 720 & 0.519 & 0.500 & 0.503 & 0.498 & 0.463 & 0.427 & 0.426 & 0.427 & 0.426 & 0.423 & 0.440 & 0.434 & 0.440 & 0.434 & 0.429 & \textbf{0.413} & 0.416 \\
\midrule
\multirow{4}{*}{\rotatebox{90}{ETTh2}} & 96 & 0.333 & 0.302 & 0.297 & 0.289 & 0.281 & 0.287 & 0.286 & 0.286 & 0.286 & 0.279 & 0.284 & 0.283 & 0.284 & 0.284 & 0.280 & \textbf{0.274} & \textbf{0.274} \\
 & 192 & 0.477 & 0.388 & 0.380 & 0.372 & 0.355 & 0.350 & 0.349 & 0.349 & 0.349 & 0.347 & 0.350 & 0.355 & 0.354 & 0.351 & 0.350 & \textbf{0.336} & 0.339 \\
 & 336 & 0.594 & 0.426 & 0.428 & 0.386 & 0.365 & 0.378 & 0.376 & 0.378 & 0.377 & 0.376 & 0.378 & 0.387 & 0.392 & 0.379 & 0.379 & \textbf{0.360} & 0.367 \\
 & 720 & 0.831 & 0.431 & 0.427 & 0.412 & 0.413 & 0.403 & 0.396 & 0.403 & 0.402 & 0.392 & 0.412 & 0.403 & 0.411 & 0.411 & 0.400 & \textbf{0.378} & 0.380 \\
\midrule
\multirow{4}{*}{\rotatebox{90}{Electricity}} & 96 & 0.197 & 0.195 & 0.148 & 0.153 & 0.141 & 0.205 & 0.154 & 0.191 & 0.152 & 0.150 & 0.158 & 0.144 & 0.155 & 0.142 & 0.139 & 0.141 & \textbf{0.127} \\
 & 192 & 0.196 & 0.199 & 0.162 & 0.166 & 0.151 & 0.220 & 0.172 & 0.205 & 0.176 & 0.172 & 0.174 & 0.166 & 0.168 & 0.160 & 0.159 & 0.158 & \textbf{0.142} \\
 & 336 & 0.209 & 0.215 & 0.178 & 0.185 & 0.173 & 0.236 & 0.203 & 0.219 & 0.197 & 0.193 & 0.191 & 0.176 & 0.185 & 0.178 & 0.176 & 0.185 & \textbf{0.165} \\
 & 720 & 0.245 & 0.256 & 0.225 & 0.225 & 0.201 & 0.270 & 0.242 & 0.252 & 0.243 & 0.234 & 0.229 & 0.207 & 0.226 & 0.210 & 0.203 & 0.224 & \textbf{0.192} \\
\midrule
\multirow{4}{*}{\rotatebox{90}{Weather}} & 96 & 0.196 & 0.177 & 0.174 & 0.163 & 0.162 & 0.183 & 0.154 & 0.163 & 0.153 & 0.147 & 0.163 & 0.152 & 0.159 & 0.151 & 0.144 & 0.142 & \textbf{0.140} \\
 & 192 & 0.237 & 0.225 & 0.221 & 0.208 & 0.208 & 0.229 & 0.200 & 0.207 & 0.197 & 0.189 & 0.207 & 0.198 & 0.199 & 0.198 & 0.184 & 0.181 & \textbf{0.171} \\
 & 336 & 0.283 & 0.278 & 0.278 & 0.251 & 0.263 & 0.288 & 0.246 & 0.254 & 0.243 & 0.234 & 0.264 & 0.250 & 0.248 & 0.249 & 0.229 & 0.225 & \textbf{0.220} \\
 & 720 & 0.345 & 0.354 & 0.358 & 0.339 & 0.340 & 0.371 & 0.311 & 0.315 & 0.305 & 0.292 & 0.350 & 0.326 & 0.314 & 0.322 & 0.296 & \textbf{0.280} & 0.282 \\
\bottomrule
\end{tabular}
}
\end{table*}

\begin{table*}[tbp]
\centering
\caption{Full MAE results of long sequence forecasting experiments.}
\label{tab:lsf_full_mae}
\resizebox{\textwidth}{!}{
\begin{tabular}{@{}lcccccc>{\columncolor{tabhighlight}}ccccc>{\columncolor{tabhighlight}}ccccc>{\columncolor{lightgreen}}c>{\columncolor{lightgreen}}c@{}}
\toprule
 & & \multicolumn{5}{c}{Baselines} & \multicolumn{5}{c}{Moirai-small} & \multicolumn{5}{c}{Moirai-base} & \multicolumn{2}{c}{DGF} \\
\cmidrule(lr){3-7}\cmidrule(lr){8-12}\cmidrule(lr){13-17}\cmidrule(lr){18-19}
Dataset & Len. & \rotatebox{90}{DLinear\,\citeyearpar{DLinear}} & \rotatebox{90}{PatchTST\,\citeyearpar{PatchTST}} & \rotatebox{90}{iTransformer\,\citeyearpar{iTransformer}} & \rotatebox{90}{TimeMixer\,\citeyearpar{wang2024timemixer}} & \rotatebox{90}{SimpleTM\,\citeyearpar{chen2025simpletm}} & \rotatebox{90}{Moirai-small} & \rotatebox{90}{\textit{+ Full finetuning}} & \rotatebox{90}{\textit{+ Prompt tuning}} & \rotatebox{90}{\textit{+ LoRA}} & \rotatebox{90}{\textit{+ MSFT}\,\citeyearpar{MSFT}} & \rotatebox{90}{Moirai-base} & \rotatebox{90}{\textit{+ Full finetuning}} & \rotatebox{90}{\textit{+ Prompt tuning}} & \rotatebox{90}{\textit{+ LoRA}} & \rotatebox{90}{\textit{+ MSFT}\,\citeyearpar{MSFT}} & \rotatebox{90}{DGF with Moirai-small} & \rotatebox{90}{DGF with Moirai-base} \\
\midrule
\multirow{4}{*}{\rotatebox{90}{ETTm1}} & 96 & 0.372 & 0.367 & 0.368 & 0.357 & 0.361 & 0.383 & 0.345 & 0.359 & 0.344 & 0.341 & 0.360 & 0.334 & 0.355 & 0.337 & 0.335 & 0.330 & \textbf{0.322} \\
 & 192 & 0.389 & 0.385 & 0.391 & 0.381 & 0.380 & 0.402 & 0.372 & 0.380 & 0.374 & 0.367 & 0.379 & 0.361 & 0.377 & 0.359 & 0.359 & 0.358 & \textbf{0.347} \\
 & 336 & 0.413 & 0.410 & 0.420 & 0.404 & 0.404 & 0.416 & 0.393 & 0.403 & 0.394 & 0.387 & 0.394 & 0.380 & 0.393 & 0.379 & 0.378 & 0.379 & \textbf{0.365} \\
 & 720 & 0.453 & 0.439 & 0.459 & 0.441 & 0.438 & 0.437 & 0.419 & 0.423 & 0.421 & 0.414 & 0.419 & 0.409 & 0.417 & 0.407 & 0.404 & 0.410 & \textbf{0.392} \\
\midrule
\multirow{4}{*}{\rotatebox{90}{ETTm2}} & 96 & 0.292 & 0.259 & 0.264 & 0.258 & 0.257 & 0.282 & 0.251 & 0.273 & 0.250 & 0.242 & 0.271 & 0.249 & 0.271 & 0.248 & 0.246 & 0.234 & \textbf{0.229} \\
 & 192 & 0.362 & 0.302 & 0.309 & 0.299 & 0.299 & 0.318 & 0.292 & 0.309 & 0.290 & 0.281 & 0.306 & 0.288 & 0.301 & 0.289 & 0.285 & \textbf{0.271} & 0.273 \\
 & 336 & 0.427 & 0.343 & 0.348 & 0.340 & 0.338 & 0.355 & 0.329 & 0.343 & 0.327 & 0.314 & 0.339 & 0.325 & 0.339 & 0.327 & 0.320 & \textbf{0.301} & 0.308 \\
 & 720 & 0.522 & 0.400 & 0.407 & 0.396 & 0.395 & 0.410 & 0.390 & 0.409 & 0.390 & 0.368 & 0.388 & 0.367 & 0.387 & 0.366 & 0.369 & 0.361 & \textbf{0.352} \\
\midrule
\multirow{4}{*}{\rotatebox{90}{ETTh1}} & 96 & 0.400 & 0.419 & 0.405 & 0.400 & 0.392 & 0.402 & 0.400 & 0.402 & 0.399 & 0.401 & 0.398 & 0.396 & 0.397 & 0.397 & 0.395 & 0.399 & \textbf{0.384} \\
 & 192 & 0.432 & 0.445 & 0.436 & 0.421 & 0.421 & 0.423 & 0.423 & 0.423 & 0.423 & 0.421 & 0.417 & 0.416 & 0.416 & 0.416 & 0.415 & 0.415 & \textbf{0.401} \\
 & 336 & 0.459 & 0.466 & 0.458 & 0.458 & 0.438 & 0.435 & 0.438 & 0.435 & 0.435 & 0.433 & 0.429 & 0.429 & 0.428 & 0.429 & 0.429 & 0.426 & \textbf{0.418} \\
 & 720 & 0.516 & 0.488 & 0.491 & 0.482 & 0.462 & 0.450 & 0.453 & 0.451 & 0.450 & 0.449 & 0.452 & 0.454 & 0.452 & 0.451 & 0.450 & 0.440 & \textbf{0.437} \\
\midrule
\multirow{4}{*}{\rotatebox{90}{ETTh2}} & 96 & 0.387 & 0.348 & 0.349 & 0.341 & 0.338 & 0.334 & 0.332 & 0.337 & 0.333 & 0.326 & 0.334 & 0.330 & 0.335 & 0.334 & 0.327 & 0.325 & \textbf{0.320} \\
 & 192 & 0.476 & 0.400 & 0.400 & 0.392 & 0.387 & 0.374 & 0.372 & 0.373 & 0.373 & 0.369 & 0.380 & 0.378 & 0.378 & 0.378 & 0.374 & \textbf{0.357} & 0.362 \\
 & 336 & 0.541 & 0.433 & 0.432 & 0.414 & 0.401 & 0.395 & 0.392 & 0.394 & 0.394 & 0.391 & 0.405 & 0.402 & 0.404 & 0.405 & 0.404 & \textbf{0.389} & 0.398 \\
 & 720 & 0.657 & 0.446 & 0.445 & 0.434 & 0.436 & 0.421 & 0.416 & 0.420 & 0.420 & 0.413 & 0.432 & 0.429 & 0.432 & 0.431 & 0.427 & \textbf{0.401} & 0.415 \\
\midrule
\multirow{4}{*}{\rotatebox{90}{Electricity}} & 96 & 0.282 & 0.285 & 0.240 & 0.247 & 0.235 & 0.299 & 0.242 & 0.285 & 0.244 & 0.241 & 0.248 & 0.236 & 0.246 & 0.234 & 0.230 & 0.232 & \textbf{0.219} \\
 & 192 & 0.285 & 0.289 & 0.253 & 0.256 & 0.247 & 0.310 & 0.265 & 0.294 & 0.266 & 0.262 & 0.263 & 0.256 & 0.258 & 0.252 & 0.252 & 0.252 & \textbf{0.239} \\
 & 336 & 0.301 & 0.305 & 0.269 & 0.277 & 0.267 & 0.323 & 0.289 & 0.307 & 0.285 & 0.282 & 0.278 & 0.267 & 0.274 & 0.269 & 0.266 & 0.268 & \textbf{0.251} \\
 & 720 & 0.333 & 0.337 & 0.317 & 0.310 & 0.293 & 0.347 & 0.319 & 0.331 & 0.321 & 0.316 & 0.307 & 0.295 & 0.305 & 0.296 & 0.293 & 0.307 & \textbf{0.274} \\
\midrule
\multirow{4}{*}{\rotatebox{90}{Weather}} & 96 & 0.255 & 0.218 & 0.214 & 0.209 & 0.207 & 0.216 & 0.189 & 0.199 & 0.189 & 0.185 & 0.198 & 0.186 & 0.196 & 0.186 & 0.182 & 0.177 & \textbf{0.167} \\
 & 192 & 0.296 & 0.260 & 0.254 & 0.250 & 0.248 & 0.258 & 0.236 & 0.241 & 0.233 & 0.229 & 0.240 & 0.235 & 0.235 & 0.235 & 0.224 & 0.215 & \textbf{0.214} \\
 & 336 & 0.335 & 0.297 & 0.296 & 0.287 & 0.290 & 0.297 & 0.272 & 0.276 & 0.271 & 0.266 & 0.282 & 0.278 & 0.272 & 0.278 & 0.261 & 0.253 & \textbf{0.251} \\
 & 720 & 0.381 & 0.348 & 0.349 & 0.341 & 0.341 & 0.346 & 0.317 & 0.317 & 0.315 & 0.311 & 0.338 & 0.333 & 0.318 & 0.342 & 0.311 & 0.297 & \textbf{0.295} \\
\bottomrule
\end{tabular}
}
\end{table*}

\begin{table*}[tbp]
\centering
\caption{Full results for probabilistic forecasting experiments.}
\label{tab:pf_full}
\resizebox{\textwidth}{!}{
\begin{tabular}{@{}lcccc>{\columncolor{tabhighlight}}ccccc>{\columncolor{tabhighlight}}ccccc>{\columncolor{lightgreen}}c>{\columncolor{lightgreen}}c@{}}
\toprule
 & & \multicolumn{3}{c}{Baselines} & \multicolumn{5}{c}{Moirai-small} & \multicolumn{5}{c}{Moirai-base} & \multicolumn{2}{c}{DGF} \\
\cmidrule(lr){3-5}\cmidrule(lr){6-10}\cmidrule(lr){11-15}\cmidrule(lr){16-17}
Dataset & Metric & \rotatebox{90}{DeepAR\,\citeyearpar{salinas2020deepar}} & \rotatebox{90}{PatchTST\,\citeyearpar{PatchTST}} & \rotatebox{90}{TiDE\,\citeyearpar{TiDE}} & \rotatebox{90}{Moirai-small} & \rotatebox{90}{\textit{+ Full finetuning}} & \rotatebox{90}{\textit{+ Prompt tuning}} & \rotatebox{90}{\textit{+ LoRA}} & \rotatebox{90}{\textit{+ MSFT}\,\citeyearpar{MSFT}} & \rotatebox{90}{Moirai-base} & \rotatebox{90}{\textit{+ Full finetuning}} & \rotatebox{90}{\textit{+ Prompt tuning}} & \rotatebox{90}{\textit{+ LoRA}} & \rotatebox{90}{\textit{+ MSFT}\,\citeyearpar{MSFT}} & \rotatebox{90}{DGF with Moirai-small} & \rotatebox{90}{DGF with Moirai-base} \\
\midrule
\multirow{4}{*}{\shortstack{Electricity}} & sMAPE & 0.118 & 0.107 & 0.102 & 0.134 & 0.112 & 0.125 & 0.123 & 0.095 & 0.111 & 0.100 & 0.109 & 0.103 & 0.094 & 0.087 & \textbf{0.081} \\
 & MASE & 0.844 & 0.753 & 0.706 & 0.981 & 0.810 & 0.887 & 0.872 & 0.664 & 0.792 & 0.716 & 0.783 & 0.746 & 0.653 & 0.650 & \textbf{0.641} \\
 & ND & 0.080 & 0.065 & 0.061 & 0.092 & 0.070 & 0.084 & 0.079 & 0.059 & 0.069 & 0.063 & 0.069 & 0.064 & 0.058 & 0.054 & \textbf{0.051} \\
 & NRMSE & 0.704 & 0.506 & 0.514 & 0.840 & 1.260 & 0.698 & 0.650 & 0.478 & 0.551 & 0.517 & 0.583 & 0.508 & 0.471 & 0.469 & \textbf{0.459} \\
\midrule
\multirow{4}{*}{\shortstack{Solar}} & sMAPE & 1.385 & 1.501 & 1.400 & 1.445 & 1.400 & 1.413 & 1.391 & 1.381 & 1.410 & 1.282 & 1.407 & 1.387 & \textbf{1.264} & 1.354 & 1.296 \\
 & MASE & 1.222 & 1.607 & 1.265 & 1.465 & 1.181 & 1.331 & 1.160 & 1.113 & 1.292 & 0.552 & 1.285 & 1.184 & \textbf{0.422} & 1.089 & 0.459 \\
 & ND & 0.520 & 0.685 & 0.538 & 0.624 & 0.504 & 0.567 & 0.495 & 0.475 & 0.551 & 0.239 & 0.548 & 0.505 & \textbf{0.184} & 0.468 & 0.201 \\
 & NRMSE & 1.033 & 1.408 & 1.093 & 1.135 & 1.000 & 1.081 & 0.953 & 0.949 & 1.034 & 0.554 & 1.053 & 0.967 & 0.452 & 0.932 & \textbf{0.447} \\
\midrule
\multirow{4}{*}{\shortstack{Weather}} & sMAPE & 0.776 & 0.668 & 0.636 & 0.686 & 0.612 & 0.685 & 0.617 & 0.605 & 0.623 & 0.626 & 0.613 & 0.610 & 0.622 & 0.609 & \textbf{0.591} \\
 & MASE & 3.170 & 0.844 & 0.832 & 0.521 & 0.466 & 0.520 & 0.472 & 0.451 & 0.487 & 0.511 & 0.484 & 0.465 & 0.474 & \textbf{0.449} & 0.465 \\
 & ND & 0.163 & 0.072 & 0.066 & 0.063 & 0.043 & 0.063 & 0.043 & 0.043 & 0.048 & 0.045 & 0.046 & 0.043 & 0.044 & 0.042 & \textbf{0.040} \\
 & NRMSE & 0.486 & 0.260 & 0.214 & 0.229 & 0.200 & 0.232 & 0.197 & 0.198 & 0.417 & 2.980 & 0.659 & 0.717 & 0.636 & \textbf{0.192} & 0.398 \\
\midrule
\multirow{4}{*}{\shortstack{Istanbul\\Traffic}} & sMAPE & \textbf{0.249} & 0.287 & 0.280 & 0.359 & 0.319 & 0.302 & 0.327 & 0.295 & 0.284 & 0.251 & 0.288 & 0.263 & 0.289 & 0.289 & 0.281 \\
 & MASE & 0.613 & 0.653 & 0.618 & 0.990 & 0.827 & 0.815 & 0.907 & 0.815 & 0.644 & 0.620 & 0.573 & 0.621 & 0.568 & 0.801 & \textbf{0.542} \\
 & ND & 0.139 & 0.148 & 0.140 & 0.224 & 0.188 & 0.185 & 0.206 & 0.182 & 0.146 & 0.140 & 0.130 & 0.141 & 0.129 & 0.179 & \textbf{0.121} \\
 & NRMSE & 0.181 & 0.190 & 0.185 & 0.294 & 0.298 & 0.284 & 0.282 & 0.252 & 0.194 & 0.251 & 0.170 & 0.219 & 0.160 & 0.247 & \textbf{0.149} \\
\midrule
\multirow{4}{*}{\shortstack{Turkey\\Power}} & sMAPE & 0.404 & 0.416 & 0.389 & 0.389 & 0.378 & 0.387 & 0.382 & 0.377 & 0.378 & 0.372 & 0.377 & 0.371 & 0.372 & 0.373 & \textbf{0.367} \\
 & MASE & 1.395 & 1.234 & 0.904 & 0.948 & 0.863 & 0.947 & 0.887 & 0.864 & 0.888 & 0.816 & 0.866 & 0.825 & 0.814 & 0.866 & \textbf{0.810} \\
 & ND & 0.083 & 0.071 & 0.059 & 0.061 & 0.048 & 0.058 & 0.055 & 0.051 & 0.051 & 0.045 & 0.052 & 0.045 & 0.045 & 0.053 & \textbf{0.042} \\
 & NRMSE & 0.181 & 0.158 & 0.139 & 0.149 & 0.124 & 0.142 & 0.131 & 0.122 & 0.118 & 0.101 & 0.120 & 0.101 & 0.099 & 0.125 & \textbf{0.093} \\
\bottomrule
\end{tabular}}
\end{table*}

\clearpage
\newpage
\input{checklist.tex}

\end{document}

%% file: checklist.tex
\section*{NeurIPS Paper Checklist}

\begin{enumerate}

\item {\bf Claims}
    \item[] Question: Do the main claims made in the abstract and introduction accurately reflect the paper's contributions and scope?
    \item[] Answer: \answerYes{}
    \item[] Justification: The abstract and introduction state the main claims of the paper: we reinterpret over-smoothing in time series forecasting as non-convex latent dynamical mode compression under single-realization supervision, and propose DGF to preserve mode-conditioned futures through Dirichlet-guided sampling and reward-based optimization. The experimental sections evaluate forecasting accuracy, probabilistic forecast quality, over-smoothing mitigation, and component effectiveness.

\item {\bf Limitations}
    \item[] Question: Does the paper discuss the limitations of the work performed by the authors?
    \item[] Answer: \answerYes{} 
    \item[] Justification: We have discussed limitations in Appendix \ref{sec_app:discussion_impact_limitation}.

\item {\bf Theory assumptions and proofs}
    \item[] Question: For each theoretical result, does the paper provide the full set of assumptions and a complete (and correct) proof?
    \item[] Answer: \answerYes{}
    \item[] Justification: We have provided the detailed proofs and analyses in Appendix \ref{app:proofs}.

    \item {\bf Experimental result reproducibility}
    \item[] Question: Does the paper fully disclose all the information needed to reproduce the main experimental results of the paper to the extent that it affects the main claims and/or conclusions of the paper (regardless of whether the code and data are provided or not)?
    \item[] Answer: \answerYes{}
    \item[] Justification: We provide benchmark descriptions, implementation details, and evaluation protocols in Section 6 and Appendix E, with additional ablations and full results in Appendices F and G.
    

\item {\bf Open access to data and code}
    \item[] Question: Does the paper provide open access to the data and code, with sufficient instructions to faithfully reproduce the main experimental results, as described in supplemental material?
    \item[] Answer: \answerYes{}
    \item[] Justification: The implementation is available at \url{https://github.com/Zh-XY22/DGF}. Experimental settings and evaluation protocols are described in Appendices E and F.

\item {\bf Experimental setting/details}
    \item[] Question: Does the paper specify all the training and test details (e.g., data splits, hyperparameters, how they were chosen, type of optimizer) necessary to understand the results?
    \item[] Answer: \answerYes{}
    \item[] Justification: We specify datasets, evaluation protocols, finetuning baselines, optimizer, learning-rate search space, batch-size search space, DGF hyperparameters, and deterministic decision protocols in Section~\ref{exp}, Appendix~\ref{sec:app_imple_details}, and Appendix~\ref{app_ablation}.

\item {\bf Experiment statistical significance}
    \item[] Question: Does the paper report error bars suitably and correctly defined or other appropriate information about the statistical significance of the experiments?
    \item[] Answer: \answerYes{}
    \item[] Justification: We repeat the main experiments with multiple random seeds and report mean and standard deviation for the over-smoothing analysis in Table~\ref{tab:oversmoothing_stripe}.

\item {\bf Experiments compute resources}
    \item[] Question: For each experiment, does the paper provide sufficient information on the computer resources (type of compute workers, memory, time of execution) needed to reproduce the experiments?
    \item[] Answer: \answerYes{}
    \item[] Justification: We discuss in Appendix \ref{app:cost_analysis}.
    
\item {\bf Code of ethics}
    \item[] Question: Does the research conducted in the paper conform, in every respect, with the NeurIPS Code of Ethics \url{https://neurips.cc/public/EthicsGuidelines}?
    \item[] Answer: \answerYes{}
    \item[] Justification: This work conforms, in every respect, with the NeurIPS Code of Ethics.

\item {\bf Broader impacts}
    \item[] Question: Does the paper discuss both potential positive societal impacts and negative societal impacts of the work performed?
    \item[] Answer: \answerYes{}
    \item[] Justification: We have discussed this in Appendix \ref{sec_app:discussion_impact_limitation}.
    
\item {\bf Safeguards}
    \item[] Question: Does the paper describe safeguards that have been put in place for responsible release of data or models that have a high risk for misuse (e.g., pre-trained language models, image generators, or scraped datasets)?
    \item[] Answer: \answerNA{}
    \item[] Justification: The paper does not release a high-risk generative model, scraped dataset, or model intended for dual-use applications. The method is evaluated on public time series forecasting benchmarks.

\item {\bf Licenses for existing assets}
    \item[] Question: Are the creators or original owners of assets (e.g., code, data, models), used in the paper, properly credited and are the license and terms of use explicitly mentioned and properly respected?
    \item[] Answer: \answerYes{} 
    \item[] Justification: We have cited all used data and provided the licenses properly.

\item {\bf New assets}
    \item[] Question: Are new assets introduced in the paper well documented and is the documentation provided alongside the assets?
    \item[] Answer: \answerYes{}
    \item[] Justification: We provide documentation for the released implementation, including training scripts, evaluation scripts, configuration files, and instructions for reproducing the main experiments.

\item {\bf Crowdsourcing and research with human subjects}
    \item[] Question: For crowdsourcing experiments and research with human subjects, does the paper include the full text of instructions given to participants and screenshots, if applicable, as well as details about compensation (if any)? 
    \item[] Answer: \answerNA{}
    \item[] Justification: This work does not involve crowdsourcing, user studies, or human-subject research.

\item {\bf Institutional review board (IRB) approvals or equivalent for research with human subjects}
    \item[] Question: Does the paper describe potential risks incurred by study participants, whether such risks were disclosed to the subjects, and whether Institutional Review Board (IRB) approvals (or an equivalent approval/review based on the requirements of your country or institution) were obtained?
    \item[] Answer: \answerNA{} 
    \item[] Justification: This work does not involve crowdsourcing, user studies, or human-subject research, so IRB approval is not applicable.

\item {\bf Declaration of LLM usage}
    \item[] Question: Does the paper describe the usage of LLMs if it is an important, original, or non-standard component of the core methods in this research? Note that if the LLM is used only for writing, editing, or formatting purposes and does \emph{not} impact the core methodology, scientific rigor, or originality of the research, declaration is not required.
    \item[] Answer: \answerNA{} 
    \item[] Justification: We only use LLM for editing, which does not involve LLMs as any important, original, or non-standard components.

\end{enumerate}

%% file: reference.bib
@misc{TSF2ECL,
      title={Comparative Analysis of Time Series Forecasting Approaches for Household Electricity Consumption Prediction}, 
      author={Muhammad Bilal and Hyeok Kim and Muhammad Fayaz and Pravin Pawar},
      year={2022},
      eprint={2207.01019},
      archivePrefix={arXiv},
      primaryClass={cs.LG}
}

@misc{TSF2weather,
  title={Time Series Analysis in Meteorology and Climatology: An Introduction},
  author={Duchon, Claude and Hale, Robert},
  year={2012},
  publisher={John Wiley \& Sons, Ltd}
}

@article{TSF2traffic,
  title={Trend modeling for traffic time series analysis: An integrated study},
  author={Li, Li and Su, Xiaonan and Zhang, Yi and Lin, Yuetong and Li, Zhiheng},
  journal={IEEE Transactions on Intelligent Transportation Systems},
  volume={16},
  number={6},
  pages={3430--3439},
  year={2015},
  publisher={IEEE}
}

@misc{iTransformer,
      title={iTransformer: Inverted Transformers Are Effective for Time Series Forecasting}, 
      author={Yong Liu and Tengge Hu and Haoran Zhang and Haixu Wu and Shiyu Wang and Lintao Ma and Mingsheng Long},
      year={2024},
      eprint={2310.06625},
      archivePrefix={arXiv},
      primaryClass={cs.LG},
      url={https://arxiv.org/abs/2310.06625}, 
}

@misc{bai2018empirical,
      title={An Empirical Evaluation of Generic Convolutional and Recurrent Networks for Sequence Modeling}, 
      author={Shaojie Bai and J. Zico Kolter and Vladlen Koltun},
      year={2018},
      eprint={1803.01271},
      archivePrefix={arXiv},
      primaryClass={cs.LG}
}

@article{how_RNN2TSF,
   title={Recurrent Neural Networks for Time Series Forecasting: Current status and future directions},
   volume={37},
   ISSN={0169-2070},
   url={http://dx.doi.org/10.1016/j.ijforecast.2020.06.008},
   DOI={10.1016/j.ijforecast.2020.06.008},
   number={1},
   journal={International Journal of Forecasting},
   publisher={Elsevier BV},
   author={Hewamalage, Hansika and Bergmeir, Christoph and Bandara, Kasun},
   year={2021},
   month=jan, pages={388–427} }

@misc{TransformersinTS,
      title={Transformers in Time Series: A Survey}, 
      author={Qingsong Wen and Tian Zhou and Chaoli Zhang and Weiqi Chen and Ziqing Ma and Junchi Yan and Liang Sun},
      year={2023},
      eprint={2202.07125},
      archivePrefix={arXiv},
      primaryClass={cs.LG}
}

@misc{zhan2023differential,
  title={Differential Convolutional Fuzzy Time Series Forecasting},
  author={Zhan, Tianxiang and He, Yuanpeng and Deng, Yong and Li, Zhen},
  journal={IEEE Transactions on Fuzzy Systems},
  year={2023},
  publisher={IEEE}
}

@article{tang2021building,
  title={Building trend fuzzy granulation-based LSTM recurrent neural network for long-term time-series forecasting},
  author={Tang, Yuqing and Yu, Fusheng and Pedrycz, Witold and Yang, Xiyang and Wang, Jiayin and Liu, Shihu},
  journal={IEEE transactions on fuzzy systems},
  volume={30},
  number={6},
  pages={1599--1613},
  year={2021},
  publisher={IEEE}
}

@misc{TimesNet,
  title={TimesNet: Temporal 2D-Variation Modeling for General Time Series Analysis},
  author={Wu, Haixu and Hu, Tengge and Liu, Yong and Zhou, Hang and Wang, Jianmin and Long, Mingsheng},
  journal={ICLR},
  year={2023}
}

@misc{PatchTST,
  title     = {A Time Series is Worth 64 Words: Long-term Forecasting with Transformers},
  author    = {Nie, Yuqi and
               H. Nguyen, Nam and
               Sinthong, Phanwadee and 
               Kalagnanam, Jayant},
  journal = {International Conference on Learning Representations},
  year      = {2023}
}

@misc{FEDformer,
  title={{FEDformer}: Frequency enhanced decomposed transformer for long-term series forecasting},
  author={Zhou, Tian and Ma, Ziqing and Wen, Qingsong and Wang, Xue and Sun, Liang and Jin, Rong},
  journal={Proc. 39th International Conference on Machine Learning (ICML 2022)},
  address = {Baltimore, Maryland},
  pages={},
  year={2022}
}

@misc{DLinear,
  title={Are Transformers Effective for Time Series Forecasting?},
  author={Ailing Zeng and Muxi Chen and Lei Zhang and Qiang Xu},
  journal={AAAI},
  year={2023}
}

@article{Pytorch,
  title={PyTorch: An Imperative Style, High-Performance Deep Learning Library},
  author={Adam Paszke and S. Gross and Francisco Massa and A. Lerer and James Bradbury and Gregory Chanan and Trevor Killeen and Z. Lin and N. Gimelshein and L. Antiga and Alban Desmaison and Andreas K{\"o}pf and Edward Yang and Zach DeVito and Martin Raison and Alykhan Tejani and Sasank Chilamkurthy and Benoit Steiner and Lu Fang and Junjie Bai and Soumith Chintala},
  journal={NeurIPS},
  year={2019}
}

@article{Adam,
  author    = {Diederik P. Kingma and
               Jimmy Ba},
  title     = {Adam: {A} Method for Stochastic Optimization},
  journal = {ICLR},
  year      = {2015}
}

@misc{TiDE,
  title={Long-term Forecasting with TiDE: Time-series Dense Encoder},
  author={Das, Abhimanyu and Kong, Weihao and Leach, Andrew and Sen, Rajat and Yu, Rose},
  journal={arXiv preprint arXiv:2304.08424},
  year={2023}
}

@article{lecun_deep_2015,
	title = {Deep learning},
	volume = {521},
	copyright = {2015 Springer Nature Limited},
	issn = {1476-4687},
	url = {https://www.nature.com/articles/nature14539},
	doi = {10.1038/nature14539},
	language = {en},
	number = {7553},
	urldate = {2023-10-31},
	journal = {Nature},
	author = {LeCun, Yann and Bengio, Yoshua and Hinton, Geoffrey},
	month = may,
	year = {2015},
	note = {Number: 7553
Publisher: Nature Publishing Group},
	pages = {436--444},
}

@misc{wang2024timemixer,
      title={TimeMixer: Decomposable Multiscale Mixing for Time Series Forecasting}, 
      author={Shiyu Wang and Haixu Wu and Xiaoming Shi and Tengge Hu and Huakun Luo and Lintao Ma and James Y. Zhang and Jun Zhou},
      year={2024},
      eprint={2405.14616},
      archivePrefix={arXiv},
      primaryClass={cs.LG},
      url={https://arxiv.org/abs/2405.14616}, 
}

@misc{zhang2024not,
      title={Not All Frequencies Are Created Equal:Towards a Dynamic Fusion of Frequencies in Time-Series Forecasting}, 
      author={Xingyu Zhang and Siyu Zhao and Zeen Song and Huijie Guo and Jianqi Zhang and Changwen Zheng and Wenwen Qiang},
      year={2024},
      eprint={2407.12415},
      archivePrefix={arXiv},
      primaryClass={cs.LG},
      url={https://arxiv.org/abs/2407.12415}, 
}

@article{article,
author = {Castro, Manuel and Júnior, Pedro and Soriano Vargas, Aurea and Werneck, Rafael and Gonçalves, Maiara and Lusquino Filho, Leopoldo and Moura, Renato and Zampieri, Marcelo and Linares, Oscar and Ferreira, Vitor and Ferreira, Alexandre and Davolio, Alessandra and Schiozer, Denis and Rocha, Anderson},
year = {2023},
month = {07},
pages = {},
title = {Time series causal relationships discovery through feature importance and ensemble models},
volume = {13},
journal = {Scientific Reports},
doi = {10.1038/s41598-023-37929-w}
}

@article{lim2021time,
  title={Time-series forecasting with deep learning: a survey},
  author={Lim, Bryan and Zohren, Stefan},
  journal={Philosophical Transactions of the Royal Society A},
  volume={379},
  number={2194},
  pages={20200209},
  year={2021},
  publisher={The Royal Society Publishing}
}

@misc{PromptCast,
      title={PromptCast: A New Prompt-based Learning Paradigm for Time Series Forecasting}, 
      author={Hao Xue and Flora D. Salim},
      year={2023},
      eprint={2210.08964},
      archivePrefix={arXiv},
      primaryClass={stat.ME},
      url={https://arxiv.org/abs/2210.08964}, 
}

@misc{LLMTime,
      title={Large Language Models Are Zero-Shot Time Series Forecasters}, 
      author={Nate Gruver and Marc Finzi and Shikai Qiu and Andrew Gordon Wilson},
      year={2024},
      eprint={2310.07820},
      archivePrefix={arXiv},
      primaryClass={cs.LG},
      url={https://arxiv.org/abs/2310.07820}, 
}

@misc{FPT,
      title={One Fits All:Power General Time Series Analysis by Pretrained LM}, 
      author={Tian Zhou and PeiSong Niu and Xue Wang and Liang Sun and Rong Jin},
      year={2023},
      eprint={2302.11939},
      archivePrefix={arXiv},
      primaryClass={cs.LG},
      url={https://arxiv.org/abs/2302.11939}, 
}

@misc{UniTime,
      title={UniTime: A Language-Empowered Unified Model for Cross-Domain Time Series Forecasting}, 
      author={Xu Liu and Junfeng Hu and Yuan Li and Shizhe Diao and Yuxuan Liang and Bryan Hooi and Roger Zimmermann},
      year={2024},
      eprint={2310.09751},
      archivePrefix={arXiv},
      primaryClass={cs.LG},
      url={https://arxiv.org/abs/2310.09751}, 
}

@misc{Time-LLM,
      title={Time-LLM: Time Series Forecasting by Reprogramming Large Language Models}, 
      author={Ming Jin and Shiyu Wang and Lintao Ma and Zhixuan Chu and James Y. Zhang and Xiaoming Shi and Pin-Yu Chen and Yuxuan Liang and Yuan-Fang Li and Shirui Pan and Qingsong Wen},
      year={2024},
      eprint={2310.01728},
      archivePrefix={arXiv},
      primaryClass={cs.LG},
      url={https://arxiv.org/abs/2310.01728}, 
}

@misc{tempo,
      title={TEMPO: Prompt-based Generative Pre-trained Transformer for Time Series Forecasting}, 
      author={Defu Cao and Furong Jia and Sercan O Arik and Tomas Pfister and Yixiang Zheng and Wen Ye and Yan Liu},
      year={2024},
      eprint={2310.04948},
      archivePrefix={arXiv},
      primaryClass={cs.LG},
      url={https://arxiv.org/abs/2310.04948}, 
}

@inproceedings{AutoTimes,
author = {Liu, Yong and Qin, Guo and Huang, Xiangdong and Wang, Jianmin and Long, Mingsheng},
title = {AutoTimes: autoregressive time series forecasters via large language models},
year = {2025},
isbn = {9798331314385},
publisher = {Curran Associates Inc.},
address = {Red Hook, NY, USA},
booktitle = {Proceedings of the 38th International Conference on Neural Information Processing Systems},
articleno = {3882},
numpages = {31},
location = {Vancouver, BC, Canada},
series = {NIPS '24}
}

@article{salinas2020deepar,

title = {DeepAR: Probabilistic forecasting with autoregressive recurrent networks},

journal = {International Journal of Forecasting},

volume = {36},

number = {3},

pages = {1181-1191},

year = {2020},

issn = {0169-2070},

doi = {https://doi.org/10.1016/j.ijforecast.2019.07.001},

url = {https://www.sciencedirect.com/science/article/pii/S0169207019301888},

author = {David Salinas and Valentin Flunkert and Jan Gasthaus and Tim Januschowski}

}

@inproceedings{rangapuram2018deep,

author = {Rangapuram, Syama Sundar and Seeger, Matthias and Gasthaus, Jan and Stella, Lorenzo and Wang, Yuyang and Januschowski, Tim},

title = {Deep state space models for time series forecasting},

year = {2018},

publisher = {Curran Associates Inc.},

address = {Red Hook, NY, USA},

booktitle = {Proceedings of the 32nd International Conference on Neural Information Processing Systems},

pages = {7796–7805},

numpages = {10},

location = {Montr\'{e}al, Canada},

series = {NIPS'18}

}

@article{wen2017multi,
author = {Wen, Ruofeng and Torkkola, Kari and Narayanaswamy, Balakrishnan},
year = {2017},
month = {11},
pages = {},
title = {A Multi-Horizon Quantile Recurrent Forecaster},
doi = {10.48550/arXiv.1711.11053}
}

@InProceedings{gasthaus2019probabilistic,

  title = 	 {Probabilistic Forecasting with Spline Quantile Function RNNs},

  author =       {Gasthaus, Jan and Benidis, Konstantinos and Wang, Yuyang and Rangapuram, Syama Sundar and Salinas, David and Flunkert, Valentin and Januschowski, Tim},

  booktitle = 	 {Proceedings of the Twenty-Second International Conference on Artificial Intelligence and Statistics},

  pages = 	 {1901--1910},

  year = 	 {2019},

  editor = 	 {Chaudhuri, Kamalika and Sugiyama, Masashi},

  volume = 	 {89},

  series = 	 {Proceedings of Machine Learning Research},

  month = 	 {16--18 Apr},

  publisher =    {PMLR},

  url = 	 {https://proceedings.mlr.press/v89/gasthaus19a.html}

}

@inproceedings{

rasul2020multivariate,

title={Multivariate Probabilistic Time Series Forecasting via Conditioned Normalizing Flows},

author={Kashif Rasul and Abdul-Saboor Sheikh and Ingmar Schuster and Urs M Bergmann and Roland Vollgraf},

booktitle={International Conference on Learning Representations},

year={2021},

url={https://openreview.net/forum?id=WiGQBFuVRv}

}

@misc{TimeGrad,
title={Autoregressive Denoising Diffusion Models for Multivariate Probabilistic Time Series Forecasting}, 
author={Kashif Rasul and Calvin Seward and Ingmar Schuster and Roland Vollgraf},
year={2021},
eprint={2101.12072},
archivePrefix={arXiv},
primaryClass={cs.LG},
url={https://arxiv.org/abs/2101.12072}, 
}

@inproceedings{tashiro2021csdi,
author = {Tashiro, Yusuke and Song, Jiaming and Song, Yang and Ermon, Stefano},
title = {CSDI: conditional score-based diffusion models for probabilistic time series imputation},
year = {2021},
isbn = {9781713845393},
publisher = {Curran Associates Inc.},
address = {Red Hook, NY, USA},
booktitle = {Proceedings of the 35th International Conference on Neural Information Processing Systems},
articleno = {1900},
numpages = {13},
series = {NIPS '21}
}

@inproceedings{
liu2025timerxl,
title={Timer-{XL}: Long-Context Transformers for Unified Time Series Forecasting},
author={Yong Liu and Guo Qin and Xiangdong Huang and Jianmin Wang and Mingsheng Long},
booktitle={The Thirteenth International Conference on Learning Representations},
year={2025},
url={https://openreview.net/forum?id=KMCJXjlDDr}
}

@misc{deepseekmath,
      title={DeepSeekMath: Pushing the Limits of Mathematical Reasoning in Open Language Models}, 
      author={Zhihong Shao and Peiyi Wang and Qihao Zhu and Runxin Xu and Junxiao Song and Xiao Bi and Haowei Zhang and Mingchuan Zhang and Y. K. Li and Y. Wu and Daya Guo},
      year={2024},
      eprint={2402.03300},
      archivePrefix={arXiv},
      primaryClass={cs.CL},
      url={https://arxiv.org/abs/2402.03300}, 
}

@inproceedings{chen2025simpletm,
  title={SimpleTM: A Simple Baseline for Multivariate Time Series Forecasting},
  author={Chen, Hui and Luong, Viet and Mukherjee, Lopamudra and Singh, Vikas},
  booktitle={The Thirteenth International Conference on Learning Representations},
  year={2025}
}

@inproceedings{Moirai,
author = {Woo, Gerald and Liu, Chenghao and Kumar, Akshat and Xiong, Caiming and Savarese, Silvio and Sahoo, Doyen},
title = {Unified training of universal time series forecasting transformers},
year = {2024},
publisher = {JMLR.org},
booktitle = {Proceedings of the 41st International Conference on Machine Learning},
articleno = {2178},
numpages = {25},
location = {Vienna, Austria},
series = {ICML'24}
}

@inproceedings{
moiraimoe,
title={Moirai-MoE: Empowering Time Series Foundation Models with Sparse Mixture of Experts},
author={Xu Liu and Juncheng Liu and Gerald Woo and Taha Aksu and Yuxuan Liang and Roger Zimmermann and Chenghao Liu and Junnan Li and Silvio Savarese and Caiming Xiong and Doyen Sahoo},
booktitle={Forty-second International Conference on Machine Learning},
year={2025},
url={https://openreview.net/forum?id=SrEOUSyJcR}
}

@misc{TimeGPT,
      title={TimeGPT-1}, 
      author={Azul Garza and Cristian Challu and Max Mergenthaler-Canseco},
      year={2024},
      eprint={2310.03589},
      archivePrefix={arXiv},
      primaryClass={cs.LG},
      url={https://arxiv.org/abs/2310.03589}, 
}

@misc{TimesFM,
      title={A decoder-only foundation model for time-series forecasting}, 
      author={Abhimanyu Das and Weihao Kong and Rajat Sen and Yichen Zhou},
      year={2024},
      eprint={2310.10688},
      archivePrefix={arXiv},
      primaryClass={cs.CL},
      url={https://arxiv.org/abs/2310.10688}, 
}

@misc{Chronos,
      title={Chronos: Learning the Language of Time Series}, 
      author={Abdul Fatir Ansari and Lorenzo Stella and Caner Turkmen and Xiyuan Zhang and Pedro Mercado and Huibin Shen and Oleksandr Shchur and Syama Sundar Rangapuram and Sebastian Pineda Arango and Shubham Kapoor and Jasper Zschiegner and Danielle C. Maddix and Hao Wang and Michael W. Mahoney and Kari Torkkola and Andrew Gordon Wilson and Michael Bohlke-Schneider and Yuyang Wang},
      year={2024},
      eprint={2403.07815},
      archivePrefix={arXiv},
      primaryClass={cs.LG},
      url={https://arxiv.org/abs/2403.07815}, 
}

@misc{Timer,
      title={Timer: Generative Pre-trained Transformers Are Large Time Series Models}, 
      author={Yong Liu and Haoran Zhang and Chenyu Li and Xiangdong Huang and Jianmin Wang and Mingsheng Long},
      year={2024},
      eprint={2402.02368},
      archivePrefix={arXiv},
      primaryClass={cs.LG},
      url={https://arxiv.org/abs/2402.02368}, 
}

@misc{MOMENT,
      title={MOMENT: A Family of Open Time-series Foundation Models}, 
      author={Mononito Goswami and Konrad Szafer and Arjun Choudhry and Yifu Cai and Shuo Li and Artur Dubrawski},
      year={2024},
      eprint={2402.03885},
      archivePrefix={arXiv},
      primaryClass={cs.LG},
      url={https://arxiv.org/abs/2402.03885}, 
}

@inproceedings{
MSFT,
title={Multi-Scale Finetuning for Encoder-based Time Series Foundation Models},
author={Zhongzheng Qiao and Chenghao Liu and Yiming Zhang and Ming Jin and Quang Pham and Qingsong Wen and Ponnuthurai Nagaratnam Suganthan and Xudong Jiang and Savitha Ramasamy},
booktitle={The Thirty-ninth Annual Conference on Neural Information Processing Systems},
year={2026},
url={https://openreview.net/forum?id=OPOBV0zXu7}
}

@misc{hu2021lora,
      title={LoRA: Low-Rank Adaptation of Large Language Models}, 
      author={Edward J. Hu and Yelong Shen and Phillip Wallis and Zeyuan Allen-Zhu and Yuanzhi Li and Shean Wang and Lu Wang and Weizhu Chen},
      year={2021},
      eprint={2106.09685},
      archivePrefix={arXiv},
      primaryClass={cs.CL},
      url={https://arxiv.org/abs/2106.09685}, 
}

@inproceedings{STRIPE,
 author = {LE GUEN, Vincent and THOME, Nicolas},
 booktitle = {Advances in Neural Information Processing Systems},
 editor = {H. Larochelle and M. Ranzato and R. Hadsell and M.F. Balcan and H. Lin},
 pages = {4427--4440},
 publisher = {Curran Associates, Inc.},
 title = {Probabilistic Time Series Forecasting with Shape and Temporal Diversity},
 url = {https://proceedings.neurips.cc/paper_files/paper/2020/file/2f2b265625d76a6704b08093c652fd79-Paper.pdf},
 volume = {33},
 year = {2020}
}

@inproceedings{HCAN,
author = {Sun, Yanru and Xie, Zongxia and Chen, Dongyue and Eldele, Emadeldeen and Hu, Qinghua},
title = {Hierarchical classification auxiliary network for time series forecasting},
year = {2025},
isbn = {978-1-57735-897-8},
publisher = {AAAI Press},
url = {https://doi.org/10.1609/aaai.v39i19.34286},
doi = {10.1609/aaai.v39i19.34286},
booktitle = {Proceedings of the Thirty-Ninth AAAI Conference on Artificial Intelligence and Thirty-Seventh Conference on Innovative Applications of Artificial Intelligence and Fifteenth Symposium on Educational Advances in Artificial Intelligence},
articleno = {2313},
numpages = {9},
series = {AAAI'25/IAAI'25/EAAI'25}
}

@misc{DILATE,
      title={Shape and Time Distortion Loss for Training Deep Time Series Forecasting Models}, 
      author={Vincent Le Guen and Nicolas Thome},
      year={2019},
      eprint={1909.09020},
      archivePrefix={arXiv},
      primaryClass={stat.ML},
      url={https://arxiv.org/abs/1909.09020}, 
}

@misc{DeepState,
      title={Latent Matters: Learning Deep State-Space Models}, 
      author={Alexej Klushyn and Richard Kurle and Maximilian Soelch and Botond Cseke and Patrick van der Smagt},
      year={2026},
      eprint={2602.23050},
      archivePrefix={arXiv},
      primaryClass={cs.LG},
      url={https://arxiv.org/abs/2602.23050}, 
}
